# Can Large Language Models Simulate Human Responses? A Case Study of Stated Preference Experiments in the Context of Heating-related Choices

**Han Wang, Corresponding Author**

Centre for Transport Engineering and Modelling

Department of Civil and Environmental Engineering

Imperial College London, London, U.K.

han.wang19@ imperial.ac.uk

**Jacek Pawlak**

Centre for Transport Engineering and Modelling

Department of Civil and Environmental Engineering

Imperial College London, London, U.K.

jacek.pawlak@imperial.ac.uk

**Aruna Sivakumar**

Centre for Transport Engineering and Modelling

Department of Civil and Environmental Engineering

Imperial College London, London, U.K.

a.sivakumar@imperial.ac.uk

**Abstract**

Stated preference (SP) surveys are a key method to research how individuals make trade-offs in hypothetical, also futuristic, scenarios. In energy context this includes key decarbonisation enablement contexts, such as low-carbon technologies, distributed renewable energy generation, and demand-side response [1,2]. However, they tend to be costly, time-consuming, and can be affected by respondent fatigue and ethical constraints. Large language models (LLMs) have demonstrated remarkable capabilities in generating human-like textual responses, prompting growing interest in their application to survey research. This study investigates the use of LLMs to simulate consumer choices in energy-related SP surveys and explores their integration into data analysis workflows. A series of test scenarios were designed to systematically assess the simulation performance of several LLMs (LLaMA 3.1, Mistral, GPT-3.5 and DeepSeek-R1) at both individual and aggregated levels, considering contexts factors such as prompt design, in-context learning (ICL), chain-of-thought (CoT) reasoning, LLM types, integration with traditional choice models, and potential biases. Results indicate that while LLMs achieve an average accuracy surpassing random guessing and can provide a degree of respondent emulation, their performance remains insufficient to substitute original data. Cloud-based LLMs do not consistently outperform smaller local models. In this study, the reasoning model DeepSeek-R1 achieves the highest average accuracy (77%) and outperforms non-reasoning LLMs in accuracy, factor identification, and choice distribution alignment. Across models, systematic biases are observed against the gas boiler and no-retrofit options, with a preference for more energy-efficient alternatives. Local and cloud-based non-reasoning LLMs perform similarly in accuracy but exhibit differences in adherence to prompt requirements and susceptibility to social desirability biases. The findings suggest that previous SP choices are the most effective input factor, while longer prompts with additional factors and varied formats can cause LLMs to lose focus, reducing accuracy. Furthermore, the traditional mixed logit discrete choice model can support LLM prompt refinement, improving simulation accuracy and choice distribution alignments. LLMs, especially the reasoning LLM, also show potential in data analysis by indicating the significance of parameters in predictive models, thereby offering a qualitative complement to statistical models. Despite their limitations, pre-trained LLMs provide scalability and efficiency advantages, requiring minimal historical data compared to traditional survey and modelling methods.

## 1. Introduction

Surveys have long been the dominant tool for collecting data on consumer preferences and behaviours across fields such as energy, transport and marketing. They typically require participants to report status, make choices, or express opinions, which are subsequently analysed to inform policy and marketing strategies. Among the various survey methodologies used in research, stated preference (SP) surveys are particularly valuable for assessing consumer decision-making in hypothetical scenarios. These surveys present respondents with a set of systematically designed alternatives, each characterised by specific attributes, and ask them to indicate their preferred option [3]. This method offers the advantage of capturing a wide range of potential responses, including variations not readily observable in the real world. SP analysis has become a cornerstone for interring consumer sensitivity and trade-offs, and consequently for estimating willingness-to-pay or willingness-to-accept.

However, conventional survey methods are often costly and time-consuming, particularly in terms of participant recruitment and data collection. Moreover, since they include information about real individuals and their circumstances, sharing of such data is typically subject to cumbersome approval processes and may be restricted to specific purposes, types of analyses or analyst's geographic location. Large language models (LLMs) might offer an opportunity to address these challenges by simulating human-like responses, potentially reducing costs and accelerating insights. LLMs have gained significant global attention since the release of ChatGPT-3.5 in late 2022. The success of LLMs is largely attributed to their ability to generate human-like textual responses [4]. Consequently, researchers have started to investigate their potential for simulating human responses in surveys and market research [5]. However, LLMs are also known to generate hallucinations, meaning responses that are inaccurate yet superficially plausible, or biased outputs, due to limitations inherent in their training data and model structure, even when responding to fact-based questions [6]. Additionally, privacy, ethical, and security concerns remain when deploying LLMs in practice, due to input data being potentially remaining within the models' memory. To address these issues while accelerating LLM development, open-source models such as LLaMA and Mistral have become increasingly accessible through platforms like Hugging Face and Ollama. These approaches allow users to deploy LLMs locally, offering greater control over data handling, lower costs, and reduced privacy and security risks. Another family of models, called reasoning LLMs, are designed to simulate structured logical thinking, solve complex problems, and generate coherent inferences from prompts. Reasoning LLMs like ChatGPT-4o and DeepSeek-R1 represent a notable advancement, demonstrating increasing accuracy in performing advanced cognitive tasks [7]. However, the ability of local LLMs and reasoning LLMs to accurately simulate human responses remains largely unexplored. More specifically, LLMs can simulate responses that are syntactically and semantically accurate, i.e. they follow the language structure and the context of the input, but the extent to which they are truthful is not known.

The energy sector, given the urgency of the energy transition and climate change, is well-positioned to explore the integration of LLMs into survey methodologies to enhance decision-making. The UK government's clean power target for 2030 and net-zero target for 2050 present significant challenges that demand innovative approaches, rapid implementation, and cross-sector collaboration [8]. Understanding consumer behaviour is critical for achieving these goals, as it informs policy development and program design [9–12]. Survey research has played a key role in this process, offering valuable insights into how consumers engage with low-carbon technologies, distributed renewable energy generation, and demand-side response (DSR) programs [13].

Towards this end, SP surveys help researchers understand consumer preferences regarding energy-saving products and strategies, supporting the transition to net-zero energy systems [1,14]. For instance, Naghiyev et al. (2022) recruited 72 participants to investigate their stated preference for DSR. UK Transmission Operators commissioned SP surveys involving over 1000 participants to estimate consumers' willingness-to-pay for energy service improvements [15]. Despite their usefulness, traditional surveys face several challenges. First, recruiting representative samples and validating survey questions are both costly and time-consuming. The sample size typically remains below a few thousand participants, and the data collection process can take weeks or even months [9,10]. Second, survey responses can be affected by biases such as social desirability, where participants provide answers that align with perceived norms rather than their actual preferences, potentially skewing findings [16]. Furthermore, surveys typically require detailed contextual information to ensure comprehension, e.g. to introduce concepts such as demand-side response or new heating technologies. However, excessive text or technical jargon can lead to respondent fatigue, reducing response quality. Finally, ethical considerations related to participant involvement impose constraints on survey content and flexibility. These challenges highlight the need for alternative or complementary approaches. One possibility is to make use of LLM-generated responses where only realistic responses are needed (as opposed to 'real') or to improve the efficiency and reliability of survey-based research in the energy sector.

Recent studies [17] have explored the potential of using LLM-generated survey responses to supplement or even replace human inputs. Simulating human choices in SP surveys using LLMs has the potential to significantly reduce data collection time from weeks or months to mere hours. This approach also maintains scalability and could generate large response datasets at a fraction of the cost. However, the application of LLMs for response generation presents a fundamental challenge regarding response validation. Unlike data analysis where analysts can verify LLM outputs by replicating calculations, response generation lacks similar validation mechanisms unless benchmark data are available. Thus, the ability to generate truthful and empirically accurate responses by LLMs remains needed and domain specific. Existing applications of LLMs in data analysis have largely focused on instructing them to follow established statistical analysis models or interpret predefined diagrams[18], rather than leveraging their ability to infer patterns directly from their millions of parameters. This is partly due to challenges such as limited explainability, inherent randomness, and hallucinations. Nevertheless, given their ability to simulate human reasoning, LLMs may support analysis by identifying influential factors, detecting patterns and performing related tasks. One straightforward approach is to prompt LLMs to identify and explain the key factors underlying decisions. Although LLM-generated responses and explanations are far from replacing traditional models or human analysis, they present a significant direction for exploration.

This study explores the potential of LLMs to simulate consumer choices, particularly within the context of SP surveys in the energy sector. It evaluates the performance of various LLMs, including local models (LLaMA 3.1 and Mistral), cloud-based models (GPT-3.5), and a reasoning model (DeepSeek-R1). The experimental design comprises eleven prompt scenarios aimed at assessing simulation accuracy, in-context learning capability, general concept comprehension, and chain-of-thought reasoning. These concepts will be explained in subsequent sections, and these scenarios are designed to identify effective methods for simulating human responses with LLMs. A traditional mixed logit model is employed as a benchmark and to enhance the prompt design. The LLM outputs are structured into three components: choice predictions for accuracy assessment, explanations for reasoning evaluation, and identification of ignored factors in the prompts for data analysis. To conduct this evaluation, an existing SP survey dataset from the Heating survey is used, capturing UK participants' choices regarding heating technologies, retrofitting, and ownership and service models. The paper specifically addresses **four key questions:** (1) To what extent can LLMs accurately simulate individuals' choices in an energy SP survey?; (2) How do different types of LLMs perform in energy SP survey simulations?; (3) What is the optimal combination of factors to include in prompts for LLM simulations, considering individuals' previous choices, socio-demographic characteristics, attitudes and opinions?; and (4) How can LLMs be integrated with traditional choice models and support data analysis in research? The novelty of this study lies in the comprehensive evaluation of various LLMs, the assessment of simulation accuracy at both individual and aggregate levels, a systematic analysis of influential factors, and the exploration of integrating LLMs throughout the entire research process for understanding consumer preferences and behaviour. Moreover, the paper showcases a systematic methodology that can be employed for similar empirical contexts and in other domains, to understand the viability of use of LLMs for response generation.

## 2. Related work

Recent studies have demonstrated that LLMs can generate textual content highly similar to human-generated text, in terms of the structure (language syntax) and content (semantics). This capability has led researchers to explore the difference between LLM-generated and human-generated content, particularly within survey research across various domains and with reference to accuracy and truthfulness of the generated information.

A growing number of studies suggest that LLMs can perform on par with or even surpass humans in specific contexts. Aher, Arriaga & Kalai (2023) introduced the concept of the "Turing Experiment" to assess whether LLMs could replicate findings from classic economic, psycholinguistic, and social psychology experiments. These experiments include the Ultimatum Game, Garden Path sentences, Milgram Shock Experiment and Wisdom of Crowds. Their findings suggest that LLMs can simulate collective human intelligence rather than isolated individual responses. Webb, Holyoak & Lu (2023) found that GPT-3 and GPT-4 exhibit strong capabilities in analogical reasoning, often matching or exceeding human performance. Strachan et al. (2024) evaluated the theory of mind abilities of LLMs, from understanding false beliefs to interpreting indirect requests and recognising irony and faux pas. They compared the performance of LLMs (GPT and LLaMA 2) against human participants. The results showed that GPT-4 demonstrated human-level performance on most tasks, though it struggled with detecting faux pas. Luo et al. (2024) introduced a forward-looking benchmark, BrainBench, for predicting neuroscience results, showing that LLMs surpass experts in predicting experimental outcomes. However, critics have questioned whether such tests accurately capture the essence of concepts such as comprehension, reasoning, and understanding, arguing that these benchmarks provide only superficial proxies [23].

Given the 'reasoning' abilities, LLMs are increasingly studied to simulate human responses in surveys. Jansen, Jung & Salminen (2023) reviewed common tasks in survey research where LLMs can be applied, including survey instrument design, sampling, data management, data analysis, reporting and dissemination. A key area of innovation is LLMs' ability to simulate human responses, allowing researchers to conduct surveys without real participants.

In marketing, researchers are leveraging LLMs to model consumer preferences. Brand, Israeli & Ngwe (2024) demonstrated that the willingness-to-pay for products and features derived from LLMs responses, including LLaMA 3, Claude 3.5 and GPT-4o, closely align with those from human respondents. They also fine-tuned the LLMs with previous survey data from similar contexts to improve the alignment. Wang, Zhang & Zhang (2024) proposed a statistical data augmentation approach that integrates LLM-generated data with real data in conjoint analysis. They suggested that while LLM-generated data is not a direct substitute for human responses, it can serve as a valuable complement that reduces estimation error and achieves cost efficiencies in data collection. Goli & Singh (2024) investigated the ability of LLMs to emulate consumers' intertemporal preferences across various languages. Intertemporal preferences refer to choices between immediate and delayed rewards. They evaluated model performance using the proportion of later reward choices and the implied discount rate, with lower values indicating greater patience. The results showed that GPT models exhibited less patience than humans overall but demonstrated greater patience when prompted in languages with weak future tense references.

In political science, Argyle et al. (2023) employed GPT-3 to test whether LLMs could replicate human biases in U.S. political studies, such as those based on race and gender. They tested GPT-3 in three existing political studies, assessing its ability to classify free-form partisan text, predict voting behaviours, and extract association patterns between conceptual nodes in election studies. Their results suggested that GPT-3 could generate responses reflective of different demographic sub-populations within the U.S. Qu & Wang (2024) tested the performance of LLMs in simulating public opinion in diverse contexts. Findings suggested LLMs performed best in Western, English-speaking countries, especially the United States, highlighting geographical and linguistic biases.

In the energy domain, Fell (2024) examined whether LLMs could replicate findings from energy social surveys by using population-representative characteristics as prompts. His study replicated three consumer behaviour surveys, focusing on participation in peer-to-peer energy trading, preferences for multi-supplier electricity retail models, and stated uptake of energy-efficient appliances. The prompts incorporated factors such as age, gender, household income, household size, education, tenure, occupancy patterns, environmental concern, economic rationality, risk aversion, social trust, political orientation, place attachment, and the "big 5" personality traits. The findings indicated that LLM-generated results aligned closely with those of the original surveys, demonstrating the potential of LLMs in energy demand research. However, it is important to note the data leakage concern, as two of the three survey datasets had been published online.

One of the key barriers in use of LLM-based insights concerns tractability and explainability of the modelling outputs. Explainability, or interpretability, refers to the ability to present model outputs in ways that are understandable to humans [28,29]. Explainability fosters trust and transparency for users and helps researchers identify biases, refine model performance, and mitigate errors [30]. As LLMs scale in complexity, understanding techniques such as in-context learning (ICL) and chain-of-thought (CoT) prompting becomes crucial for improving model explainability and reliability [31]. ICL allow LLMs to perform new tasks by learning from a few examples provided in the input prompt, without requiring additional training or fine-tuning [32]. Meanwhile, CoT prompting encourages LLMs to generate intermediate reasoning steps before providing an answer, improving the transparency and coherence of responses [33].

LLMs still face challenges that limit their applicability in survey research despite their advantages in cost-effectiveness, scalability, and flexibility. These include risks of bias, hallucination, inappropriate responses, limited domain-specific knowledge, lack of transparency, and privacy concerns [4,19,25]. Hallucinations occur when LLMs generate responses that appear plausible but deviate from the input context, contradict factual knowledge, or conflict with previous outputs [6]. Addressing these issues is critical for the effective use of LLMs in simulating human responses in survey research.

In summary, existing research has primarily focused on the use of LLMs in the data collection phase and on evaluating their performance at the aggregate level. However, simulation accuracy at the individual level remains underexplored. Furthermore, limited attention has been given to having LLMs conduct data analysis and draw insights, particularly in terms of comparing their outputs with conventional statistical models. Notably, these gaps are apparent in studies using stated preference survey data, which offer rich information on participants and choice experiments, sometimes of unique nature, e.g. for hypothetical or futuristic scenarios.

## 3. Dataset

This study utilises data from the Heating Survey, a two-wave survey conducted between October 2023 and April 2024 in the UK [34]. Wave 1 collected the status quo of heating systems in UK households, while Wave 2, the SP survey, captured respondent stated preferences for heating technologies, retrofitting, as well as ownership and control models. This survey was designed using Qualtrics and two questionnaires were distributed by the market research company Norstat[1] to the same online panel of UK respondents.

Wave 1 consisted of six blocks of questions covering (1) Property characteristics, including floor area, age, insulation, and appliances; (2) Existing heating system, including the heat source and energy consumption; (3) Current heating costs; (4) Household attributes, such as household size, composition, income, and lifestyle factors, e.g. work from home, typical activities at selected times of day; (5) Environmental attitudes, such as perspectives on environmental issues and trust in information sources; (6) Perceptions of heating technologies, as well as potential assurances and incentives.

Wave 2 included three sets of stated preference experiments, along with follow-up questions to assess consumer preferences and personality traits. Each experiment set contained six choice scenarios, with each choice scenario offering three options characterised by five attributes. Table 1 provides an overview of these experiments, detailing the available options and their attributes. Further details on the survey methodology are provided in [34].

[1] www.norstatpanel.com

**Table 1. Stated choice (preference) experiments - summary** [35]

| | **SP1**<br>**Heat source technology** | **SP2**<br>**Property retrofit** | **SP3**<br>**Ownership and control** |
|---|---|---|---|
| Option A | Natural gas boiler | No retrofit | Full ownership |
| Option B | Hydrogen-ready boiler | Minor retrofit | Minor retrofit |
| Option C | Air-source heat pump | Major retrofit | Service-based |
| Attributes | Fixed cost<br>Operation cost<br>Support scheme<br>Maintenance visits<br>CO2 emission | Equipment cost<br>Retrofit cost<br>Nuisance duration<br>Savings<br>Support scheme | Upfront cost<br>Operation cost<br>Contract length<br>Energy pricing<br>Control and flexibility |
| Choice Scenarios | 6 | 6 | 6 |

After data cleaning and processing, the final dataset for this study includes 561 participants from Wave 2, the SP survey, along with their corresponding responses from Wave 1. All data used in this study were collected anonymously and contain no personally identifiable information. As the dataset is unpublished, there is no concern of data leakage through LLM pretraining.

## 4. Methodology

This study evaluates the simulation performance of various types of LLMs. The local LLMs include LLaMA 3.1 with 8 billion parameters and Mistral with 7 billion parameters, both deployed via the Ollama platform. The cloud-based LLM is GPT-3.5-turbo-0125 (estimated at approximately 20 billion parameters)[2], accessed through the OpenAI API. Although these models are not the most recent, they are suitable for this study given the constraints on storage and computational resources. In addition, the most recent open-source reasoning model DeepSeek-R1 is tested. This model is well known for its application of CoT reasoning in solving complex problems. Although DeepSeek-R1 is an open-source LLM that can be deployed locally, its advanced reasoning capabilities demand greater computational resources and result in longer processing times compared to non-reasoning LLMs of similar parameter sizes. Therefore, in this study, DeepSeek-R1 is accessed through the API to leverage its full capabilities while saving time. In general, the proposed approach aims to reflect typical scenarios available to an analyst and researcher in the energy space.

The three groups of simulation tests correspond to the three sets of stated choice experiments in the Heating SP survey, i.e. choice of heating appliance (gas boiler, air-source heat pump or hydrogen boiler), retrofit (none, minor, major) and ownership model (full, shared, service-based). Within each test group, individualised prompts are constructed with eleven scenarios of factors extracted from the individual's response to the survey. To ensure clarity and avoid potential biases from multi-turn interactions, the LLM processes only a single prompt per participant per simulation, excluding information from previous dialogues. Although studies have highlighted the value of using previous survey data from similar contexts to fine-tune LLMs or guide prompt design in marketing (Brand, Israeli & Ngwe, 2024; Wang, Zhang & Zhang, 2024), this study leaves such approaches for future research. Figure 1 presents the experimental flowchart. The prompt design and experimental setup are detailed in the following sections.

---

[2] https://community.openai.com/t/how-many-parameters-does-gpt-3-5-have/648417?utm_source=chatgpt.com

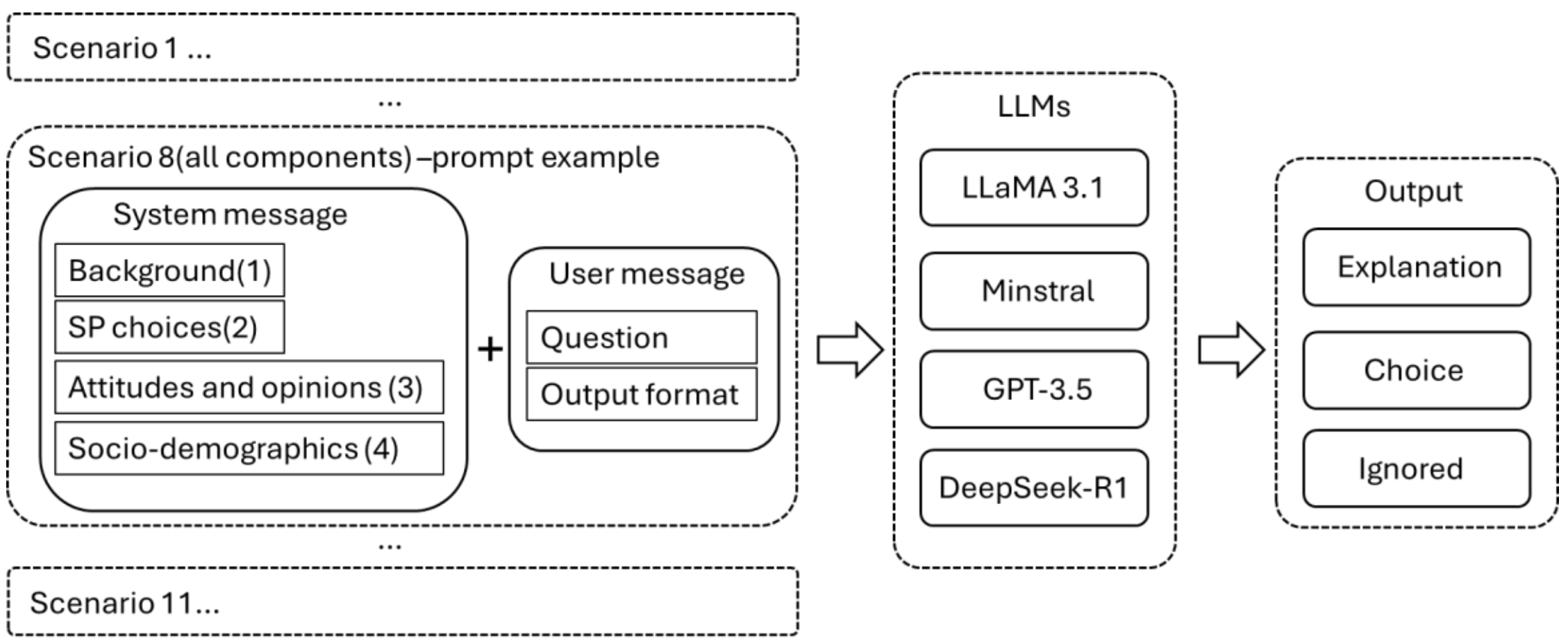

**Figure 1. Experiment flowchart**

### 4.1 Prompt

The LLM prompt usually consists of two parts, the system message and the user message. The system message provides foundational instructions, defining the LLM's role, objectives, and behavioural constraints. The user message presents the specific task or query, guiding the LLM's response. Together, this dual-message structure enhances the LLM's ability to generate accurate and contextually relevant outputs.

The system message in this study comprises up to four components depending on the test scenarios.

**System message component 1: Background**. The first component instructs the LLM to assume the identity of a UK resident living in a property with a specified built year and floor area. These property characteristics correspond to the attribute levels in the SP choice experiments and were extracted from participants' responses in Wave 1 of the Heating Survey. Sequentially this component describes one of the three sets of SP choice experiments, using sentences that closely mirror the original survey description. In certain scenarios, additional explanations of heating options, such as gas boilers and heat pumps, are included for clarity.

**System message component 2: SP choices**. The second component provides the individuals' choices in the first five choice scenarios in the same choice experiment set. The final, sixth scenario is excluded from the system message and instead presented in the user message as the simulation question. While the original Heating SP Survey displayed choice scenarios in tables for ease of comparison, the system message presents both choice scenarios and participants' choices in JSON format to align with LLM input requirements.

**System message component 3: Attitudes and opinions**. The third component includes the individuals' statements related to personality and attitudes. This information is also extracted from the Heating Survey and contains participants' responses to environmental-PVQ (E-PVQ) questions, attitudes towards replacing heating systems, and eco-friendly lifestyles. The E-PVQ metric [36] categorises environmental values into four dimensions — biospheric, altruistic, hedonic, and egoistic — to measure human values underlying environmental behaviours and beliefs. Personality and attitudinal statements used in the system message are provided in Appendix A.

**System message component 4: Socio-demographics**. The fourth component is individuals' socio-demographics, including age groups, income levels, education levels and property types. These factors are included based on prior research identifying them as key determinants of heating technology adoption and energy demand [37,38].

The user message in this study instructs LLM to select an option in the final (sixth) stated preference scenario within the same choice experiment set in the Heating SP Survey. Similarly, the options and attributes are provided in the JSON format, ensuring consistency with the system message. The LLM is explicitly required to consider the provided contextual factors when making a decision.

The user message specifies that the LLM's output must be in valid JSON format and include up to three key objects to ensure a structured and interpretable response. The first JSON object, called "Explanation", is a brief (<50 words) justification for the chosen option. The second JSON object, called "Choice", specifies the LLM's choice among three options in numerical form, e.g., 1, 2 or 3. It is also explicitly required that the LLM's decision must be consistent with the explanation. The third JSON object, called "Ignored", is in a list format and presents any factor groups in the system message that were disregarded in the LLM's decision-making process. The expected output follows this format:

{

"Explanation": "Explanation text here",

"Choice": 1 *or* 2 *or* 3,

"Ignored": ["factor1", "factor2"]

}.

This output structure captures the LLM's simulated choices, the underlying reasoning, potential sources of biases, and the significance of factors. There are several ways to utilise CoT reasoning to enhance LLM performance, including multi-round conversations (Wang, Zhang & Zhang, 2024) and explicit reasoning explanations [24]. Due to computational constraints and the number of test scenarios in this study, CoT reasoning is applied by requiring LLMs to explain reasoning before generating a choice. In traditional data analysis, models such as discrete choice models generate factor coefficients to quantify variable impacts. While asking LLMs to generate similar numerical coefficients may currently be beyond their capability, this study instead prompts them to identify ignored factors. This approach supports an exploratory integration of LLMs into the data analysis by assessing influential factors and highlighting potential sources of bias or inaccuracy. These insights can support researchers in identifying key factors for survey design and modelling.

Table 2 demonstrates the components of the prompt. Appendix A presents three example prompts, each corresponding to one of the stated choice experiments in the Heating SP Survey and containing all components.

**Table 2. Prompt component**

| Message type | Component name (number) | Content |
|---|---|---|
| System Message | Background (1) | Property characteristics |
| | | Introduction to SP choice experiment |
| | | Introduction to options and attributes |
| | SP choices in JSON format (2) | Options and their attribute levels in the first five choice scenarios |
| | | Choices in the first five choice scenarios |
| | Attitudes and opinions (3) | Environmental-PVQ |
| | | Replacing heating systems, and |
| | | Eco-friendly lifestyles |
| | Socio-demographics (4) | Age group |
| | | Income level |
| | | Education level |
| | | Property types and characteristics |
| | | Primary source of heating and heating characteristics |
| | | Location |
| | | Electricity tariff type |
| User Message | Question | Attribute levels in the 6th choice scenarios |
| | | Query to simulate the choice |
| | Output format | JSON component 1: Explanation |
| | | JSON component 2: Coded choice |
| | | JSON component 3: Ignored factors |

### 4.2 Experimental setup

This study utilises ablation analysis [25,27] , a technique that systematically adds or removes components to examine changes in model performance, to assess the contribution of various factors in LLM-based survey response simulation. The system message includes three groups of factors: SP choices, socio-demographics, attitudes and opinions. Additionally, this study examines the in-context learning (ICL) and chain of thought (CoT) reasoning abilities of LLMs and compares their performance to that of a traditional mixed multinomial logit model (MMNL). To achieve these objectives, eleven test scenarios were designed for the LLM simulation, as shown in Table 3. Prediction results from a MMNL[35] are presented as the baseline for comparison. The MMNL offers a flexible framework for capturing heterogeneity in consumer preferences [39,40].

This study evaluates four different LLMs: two local open-source models (LLaMA 3.1 and Mistral), one cloud-based model (GPT-3.5), and one reasoning-focused model (DeepSeek-R,1 hereinafter DS-R1). The parameters Temperature and Top_p are both set to 1 to encourage creative results resembling real human decision-making [19,33].

**Table 3. Test scenarios and components in prompts**

| ID | Scenario code | SP choices (SPC) | Socio-demographic (SD) | Attitudes and opinions (AO) | Removed from Scenario 8 |
|---|---|---|---|---|---|
| 1 | N (default) | | | | |
| 2 | SPC | √ | | | |
| 3 | SD | | √ | | |
| 4 | AO | | | √ | |
| 5 | SPC+SD | √ | √ | | |
| 6 | SD+AO | | √ | √ | |
| 7 | SPC+AO | √ | | √ | |
| 8 | SPC+SD+AO | √ | √ | √ | |
| 9 | SPC+SD+AO (-OP) | √ | √ | √ | Option explanations |
| 10 | SPC+SD+AO (MMNL) | √ | √ | √ | Insignificant attributes in MMNL |
| 11 | SPC+SD+AO (-CoT) | √ | √ | √ | Explanation requirements in output |

The comparisons between test scenarios 1–8 explore the optimal factor set for simulating human responses in the Heating SP Survey dataset. Scenario 9 removes the explanations of options in the system message from Scenario 8, such as the descriptions of hydrogen-ready boilers and air-source heat pumps. This scenario aims to evaluate whether LLMs can understand specific energy concepts without detailed explanations. Scenario 10 removes the statistically insignificant factors in the system message from Scenario 8 based on prior MMNL estimations [35]. This scenario explores whether traditional choice models can refine LLM prompts and improve simulation accuracy. Scenario 11 modifies the user message by removing the requirement for the "explanation" component in the output. This scenario aims to assess the impact of the CoT instruction on simulation accuracy. It is worth noting that testing DeesSeek-R1 with Scenario 11 is unnecessary, as this model is well-known for its robust and obligatory CoT capabilities. Across Scenarios 2, 5, and 7-11, the ICL ability of LLMs is tested, as previous SP choices (SPC) are provided exclusively within the prompts.

This study uses accuracy (Acc) [41] and F1 scores [42] to evaluate the simulation performance of LLMs at the individual level and determine the optimal factor set. Accuracy is the ratio of correct predictions to total number of predictions. The 95% confidence intervals for the accuracy were estimated using the bootstrap method with 20,000 resamples. The F1 score is the harmonic mean of precision and recall, providing a balanced measure of predictive performance. Precision represents the ratio of correctly predicted positive observations to the total predicted positive observations and recall measures the ratio of correctly predicted positive observations to all actual positive observations. Furthermore, the ignored component in the LLM's JSON output is analysed to understand how LLMs assess the importance of various factors.

Beyond individual-level accuracy, this study also analyses the choice distribution and stratified simulation at the aggregated level to identify potential biases. The choice distributions generated by LLMs are compared to assess simulated consumer preferences. A chi-square goodness-of-fit test was conducted to assess whether the LLM-simulated choice distribution differs significantly from the Heating SP Survey data. Additionally, the Jensen–Shannon distance was calculated to quantify the magnitude of the distributional difference. Finally, simulation results are stratified by the age groups of household members and the income level to evaluate simulation performance across various demographic segments.

## 5. Results

Approximately 5% of outputs generated by local LLMs fail to meet the valid JSON format requirement, despite explicit instructions in the user message. In contrast, all outputs from cloud-based LLMs comply with the specified format. While non-reasoning LLMs typically complete response generation within one minute per prompt, DeepSeek-R1 exhibits substantially longer processing times—approximately five-fold longer—attributable to its CoT reasoning methodology. Among the three output components, the choice component is used to calculate simulation accuracy, while the ignored factors are analysed to understand the decision-making process of LLMs. This study further assesses the impact of the presence of an explanation component on accuracy, but does not analyse the content of the explanations, as detailed text analysis lies beyond its scope.

### 4.1 Individual prediction accuracy

Table 4 summarises the prediction results of all four LLMs across eleven test scenarios, with overall accuracy and average F1 scores over three SP experiments (SP1, SP2, and SP3). Detailed results for each SP experiment are provided in Appendix B.1. For comparison, the prior mixed logit choice models achieved an accuracy of 0.60 and an F1 score of 0.52. Figure 2 illustrates the overall accuracy of all models with 95% confidence interval, using the mixed logit model and the random selection as baselines.

**Table 4. Averaged results of test scenarios**

| Scenario | | LLaMA 3.1 | | Mistral | | GPT-3.5 | | DS-R1 | |
|---|---|---|---|---|---|---|---|---|---|
| | | Acc | F1 | Acc | F1 | Acc | F1 | Acc | F1 |
| 1 | N (default) | 0.42 | 0.24 | 0.41 | 0.24 | 0.37 | 0.27 | 0.47 | 0.30 |
| 2 | SPC | 0.48 | 0.31 | 0.50 | 0.32 | 0.50 | 0.40 | 0.77 | 0.68 |
| 3 | SD | 0.41 | 0.26 | 0.40 | 0.25 | 0.39 | 0.27 | 0.47 | 0.36 |
| 4 | AO | 0.31 | 0.20 | 0.27 | 0.17 | 0.32 | 0.25 | 0.43 | 0.30 |
| 5 | SPC+SD | 0.42 | 0.28 | 0.45 | 0.28 | 0.51 | 0.39 | 0.75 | 0.72 |
| 6 | SD+AO | 0.34 | 0.22 | 0.36 | 0.24 | 0.38 | 0.28 | 0.44 | 0.35 |
| 7 | SPC+AO | 0.39 | 0.24 | 0.42 | 0.27 | 0.43 | 0.35 | 0.70 | 0.67 |
| 8 | SPC+SD+AO | 0.38 | 0.25 | 0.44 | 0.29 | 0.47 | 0.37 | 0.72 | 0.69 |
| 9 | SPC+SD+AO (-OP) | 0.39 | 0.26 | 0.44 | 0.30 | 0.51 | 0.41 | 0.72 | 0.69 |
| 10 | SPC+SD+AO (MMNL) | 0.45 | 0.30 | 0.42 | 0.27 | 0.48 | 0.35 | 0.74 | 0.72 |
| 11 | SPC+SD+AO (-CoT) | 0.38 | 0.19 | 0.44 | 0.28 | 0.38 | 0.30 | N/A | N/A |

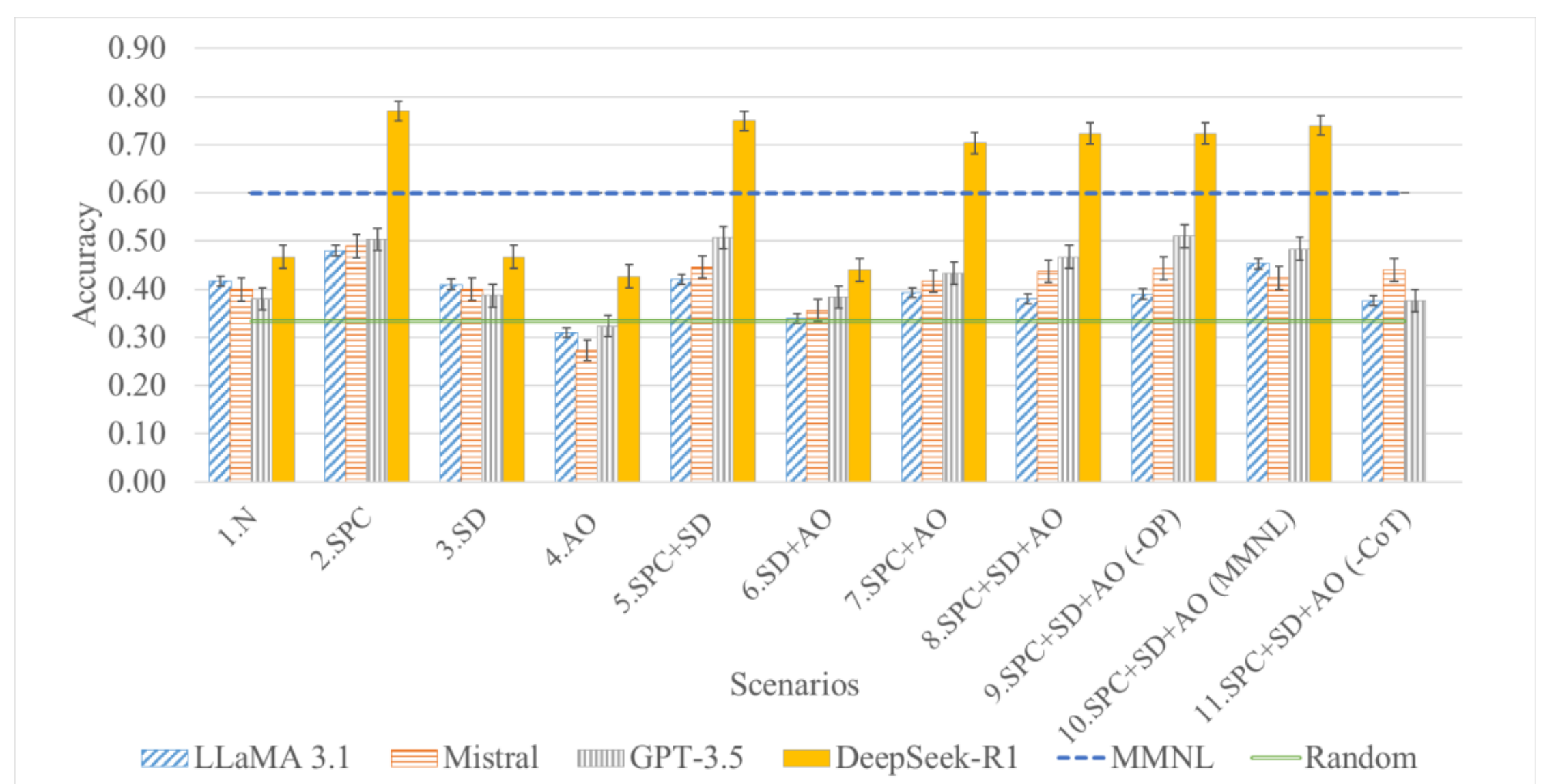

**Figure 2. Simulation accuracy of LLMs within the proposed scenarios**

Most test scenarios achieve accuracy levels higher than random guessing (Acc=0.33), demonstrating that LLM can make informed predictions based on the provided prompts. Across Scenarios 1-8, the exclusive inclusion of previous SP questions and answers (Scenario 2 SPC) results in the highest accuracy and F1 score, indicating the LLMs' In-Context Learning (ICL) capabilities in effectively inferring consumer preferences from historical choices. However, Scenarios 5-8, which introduce additional textual inputs or tokens, reduce simulation accuracy for most models, likely due to the LLM's limited ability to process long inputs efficiently.

The remaining scenarios are compared to Scenario 8 to further explore LLM capabilities. Scenario 9, which removes option explanations, does not result in lower simulation accuracy, suggesting that LLMs incorporate an inherent understanding of certain energy-related concepts. In Scenario 10, where only significant factors selected by the mixed logit model (MMNL) are included, accuracy improves for LLaMA 3.1 and DeepSeek-R1. This demonstrates the potential of integrating traditional choice models with LLM-based simulation, with selective factor inclusion in the prompt improving performance. In Scenario 11, removing the requirement for explanatory reasoning in outputs only reduces the accuracy of GPT-3.5 compared to Scenario 8, suggesting that the Chain-of-Thought (CoT) approach does not enhance predictive performance for all LLMs in this case. A possible explanation is that not all LLMs rely on reasoning processes to solve complex decision-making problems.

Across all scenarios, the cloud-based LLM, GPT-3.5-turbo-0125, slightly outperforms the local LLMs (LLaMA 3.1 and Mistral). This suggests that the trade-off between the higher accuracy of large, closed-source LLMs and the safety and privacy of smaller, open-source LLMs is worth considering in similar energy-related simulation tasks. DeepSeek-R1 stands out by outperforming both the traditional mixed logit model and other LLMs, especially when previous SP answers are included in the prompt. These results highlight the importance of inherent reasoning ability, instead of CoT requirement in prompts, and the value of prior SP responses. Although the traditional MMNL model achieves higher accuracy than most LLM-based simulations, certain LLMs demonstrate comparable performance in several cases, such as SP3 (see Appendix B.1). However, it is important to note that the MMNL model was trained on the complete dataset in a previous study to understand consumer preference [34], creating data leakage and an unfair advantage over LLMs in this comparison. Despite these limitations, these comparisons underscore the potential of LLMs as alternative predictive tools in simulating choices in SP surveys, especially in contexts with no prior data available and where realistic responses have more value than randomly generated responses, e.g. as inputs to simulations, for testing purposes or as prior distributions in SP design.

Figure 3 presents the simulation accuracy, grouped by LLMs and SP choice experiments. These three SP choice experiments vary in their real-world prevalence: SP2, concerning retrofitting, represents a common scenario; SP1, involving hydrogen boilers, is less common; and SP3 is related to emerging sharing and service-based models combined with piloting DSR programmes. The box plot shows no significant difference in accuracy across the three SP choice experiment overall. However, the accuracy for SP3 is low for LLaMA 3.1 and exhibits greater fluctuation for the other models.

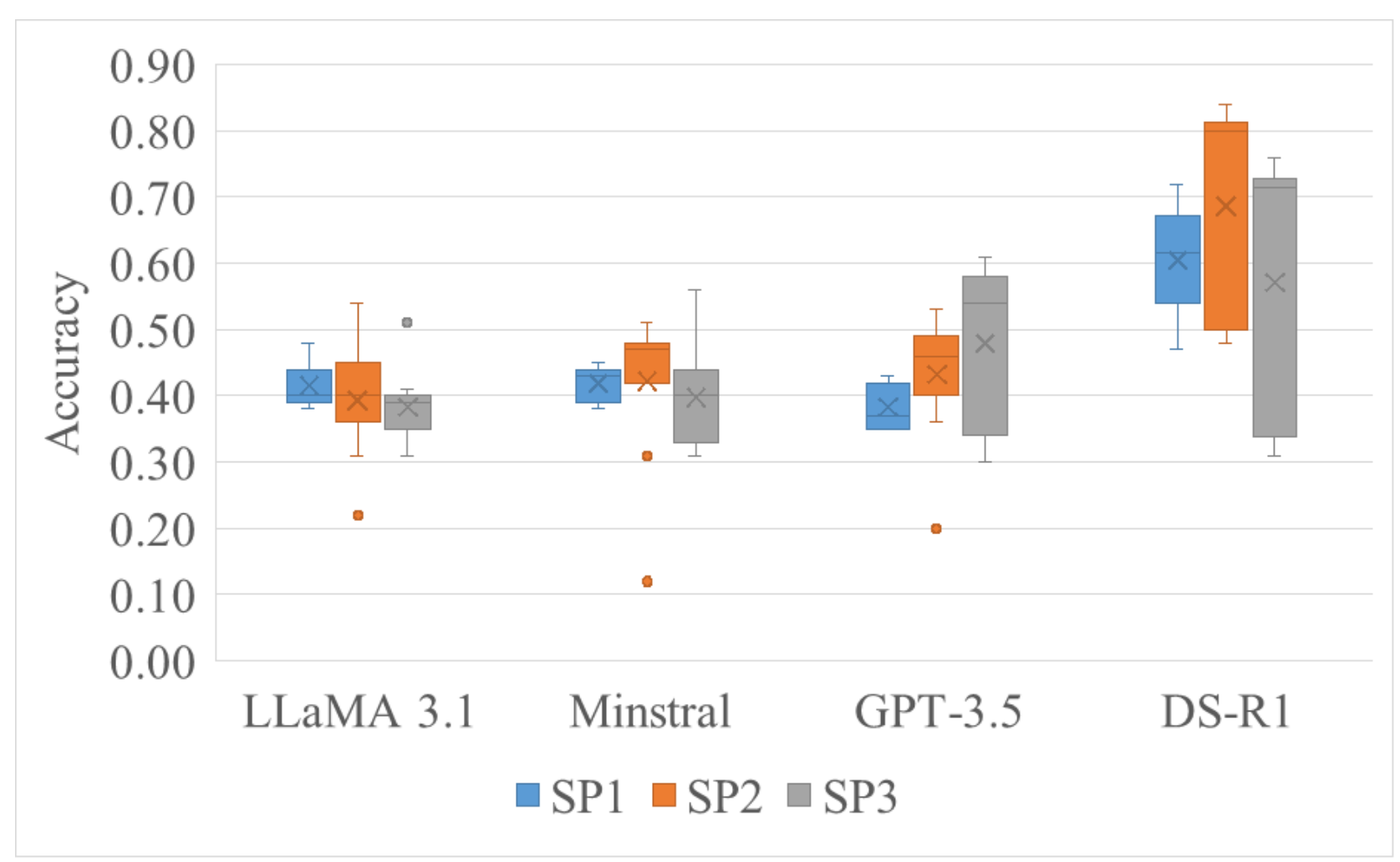

**Figure 3. Simulation accuracy of LLMs within the proposed scenarios**

### 4.2 Ignored factors

This study required LLMs to provide a list of ignored factors during the simulation as part of the output JSON format. These outputs help explain which factors does not influence the LLMs' decision-making processes, thereby exploring whether LLMs can contribute not only to data collection but also to data analysis in research. This subsection analyses the ignored factor lists in Scenarios 8-11, as their prompt includes all three types of factors, i.e., previous SP choices, socio-demographics, attitudes and opinions. Due to the inherent randomness of LLM responses, the ignored factor lists vary in granularity, ranging from broad factor types to specific attributes and socio-demographic variables. Notably, differences exist between the ignored factor lists generated by non-reasoning LLMs (LLaMA 3.1, Mistral and GPT-3.5) and the reasoning LLM (DeepSeek-R1).

For non-reasoning LLMs, the three most frequently ignored factor types, i.e., Previous SP choices (SPC), Socio-demographic (SD) and Attitudes and opinions (AO), are reported in Appendix B.2. The results suggest that previous SP choices (SPC) are the most frequently ignored factors across Scenarios 8–12 for all non-reasoning LLMs. However, as shown in Section 4.1, LLMs in Scenario 2, which includes only SPC factors, outperforms those in Scenarios 8–12 in prediction accuracy. This suggests that ignoring SPC factors may lead to lower simulation accuracy in Scenarios 8–12. This pattern indicates that LLMs may have difficulty handling long textual prompts and identifying relevant factors within them. In addition, the distinct JSON formatting of SPC factors, compared to the text-based format of other factors, may contribute to their omission in simulation. One potential reason is that information in different formats may receive different weights in attention mechanisms, and this format inconsistency increases the cognitive load for LLMs.

The reasoning LLM, DeepSeek-R1, generated distinct different ignored factor lists. The most frequently ignored factors (with frequency > 100 out of 561 survey participants) in each SP experiment simulation are shown in Table 5. The row labelled "MMNL" indicates whether these factors are significant (***) or not significant (NS) in a traditional mixed logit choice model [35], or not applicable (N/A). These ignored factors include options (e.g., major retrofit in SP2 and service-based ownership in SP3), attributes (e.g., maintenance visits, support scheme, $CO_2$ emissions, savings, contract length, and energy pricing), socio-demographics (e.g., EPC rating of properties), attitudes and opinions (e.g., agreement to prevent environmental pollution), and even specific attribute details (e.g., compensation for energy flexibility pricing). The reasoning model identified a broader and more varied set of factors than other LLMs and those typically considered in traditional choice models. Although the factors ignored by DeepSeek-R1 partially correspond to those found insignificant in the traditional MMNL model, the alignment between methods is incomplete, and the relative accuracy of these findings cannot be determined. This is because these ignored factors are identified at the individual level, while factor significance in MMNL is generalised across the sample. Nonetheless, the reasoning model demonstrates the ability to present the potential significance of factors that warrant further investigation in future research.

**Table 5. Percentage of simulated responses generated by DeepSeek-R1 containing ignored factors**

| Scenario | SP1 | | | |
| --- | --- | --- | --- | --- |
| ID | Maintenance visits | Support scheme | $CO_2$ emission | EPC rating |
| 8 | 70.23% | 27.63% | 27.63% | 26.20% |
| 9 | 62.03% | 27.81% | 27.63% | 30.66% |
| 10 | 62.57% | 19.07% | 21.75% | 65.06% |
| MMNL | NS | *** | NS | *** |

| | SP2 | | | | | |
| --- | --- | --- | --- | --- | --- | --- |
| | Prevent environmental pollution | Savings | AO | Major retrofit | Support scheme | EPC rating |
| 8 | 39.75% | 45.99% | 44.74% | 31.37% | 28.88% | 28.70% |
| 9 | 40.46% | 41.53% | 41.35% | 29.95% | 40.82% | 30.12% |
| 10 | 12.30% | 15.86% | 30.12% | 10.34% | 53.65% | 0.00% |
| MMNL | *** | NS | N/A | N/A | NS | NS |

| | SP3 | | | | | |
| --- | --- | --- | --- | --- | --- | --- |
| | Prevent environmental pollution | Energy pricing | Contract length | AO | Service-based | Compensation |
| 8 | 32.26% | 33.87% | 36.54% | 38.68% | 29.77% | 22.82% |
| 9 | 37.97% | 32.26% | 34.05% | 39.57% | 24.60% | 23.17% |
| 10 | 20.50% | 31.55% | 39.57% | 26.02% | 14.26% | 31.02% |
| MMNL | *** | *** | *** | N/A | N/A | N/A |

### 4.3 Aggregated choice distribution

The distribution of recorded choices in the Heating SP survey and simulated choices within each choice experiment is demonstrated in Appendix B.3. The meaning of options A, B and C are provided in Section 3 Table 1.

In choice experiment SP1, option A (gas boiler) is generally considered less eco-friendly than option B (Hydrogen-ready boiler) and option C (air-source heat pump).

Similarly, in choice experiment SP2, option B (minor retrofit) and option C (major retrofit) as expected to improve the energy efficiency than option A (no retrofit). Appendix B.3 reveals a systematic biases against the gas boiler and no retrofit options in simulated choices across all scenarios, particularly in outputs generated by GPT-3.5. This bias may arise from the fact that LLMs are trained to align closely with socially desirable traits [43]. Such bias could pose a significant challenge for accurately simulating human responses with LLMs.

Table 6 reports the Jensen–Shannon distance of the simulated choice distribution, averaged across three SP experiments, with smaller values indicating a closer match to the recorded distribution. Figure 4 illustrates the variation in Jensen–Shannon distance for each LLM across scenarios, using the MMNL as the baseline. Chi-square goodness-of-fit tests show that all simulated choice distributions differ significantly from the recorded choice distribution ($p < 0.05$, 95% confidence level). However, results from LLaMA 3.1 and DeepSeek-R1 show closer alignment (i.e., lower Jensen–Shannon distance) with the recorded choice distribution, particularly when previous SP answers and factors selected by the mixed logit model are included in the prompt (Scenarios 2, 5 and 10). This finding suggests that integrating traditional choice models into the prompt design may help mitigate social desirability biases in LLM-based simulations. Future research could examine whether the benefits of integrating traditional choice models persist when LLMs are used to simulate responses to adjusted questions or entirely new contexts with different datasets. Notably, consumer preferences recorded in the SP survey and derived from choice models also exhibit social desirability biases [16]. Therefore, real-world observations are necessary for further bias mitigation.

**Table 6. Average Jensen–Shannon distance of simulated choice distribution**

| | | LLaMA 3.1 | Mistral | GPT-3.5 | DeepSeek-R1 |
|---|---|---|---|---|---|
| 1 | N (default) | 0.30 | 0.26 | 0.44 | 0.46 |
| 2 | SPC | 0.20 | 0.31 | 0.35 | 0.10 |
| 3 | SD | 0.23 | 0.23 | 0.45 | 0.38 |
| 4 | AO | 0.37 | 0.53 | 0.45 | 0.40 |
| 5 | SPC+SD | 0.16 | 0.29 | 0.35 | 0.12 |
| 6 | SD+AO | 0.32 | 0.31 | 0.41 | 0.37 |
| 7 | SPC+AO | 0.28 | 0.36 | 0.39 | 0.15 |
| 8 | SPC+SD+AO | 0.27 | 0.30 | 0.38 | 0.11 |
| 9 | SPC+SD+AO (-OP) | 0.27 | 0.30 | 0.34 | 0.10 |
| 10 | SPC+SD+AO (MMNL) | 0.16 | 0.37 | 0.38 | 0.11 |
| 11 | SPC+SD+AO (-CoT) | 0.35 | 0.33 | 0.27 | N/A |

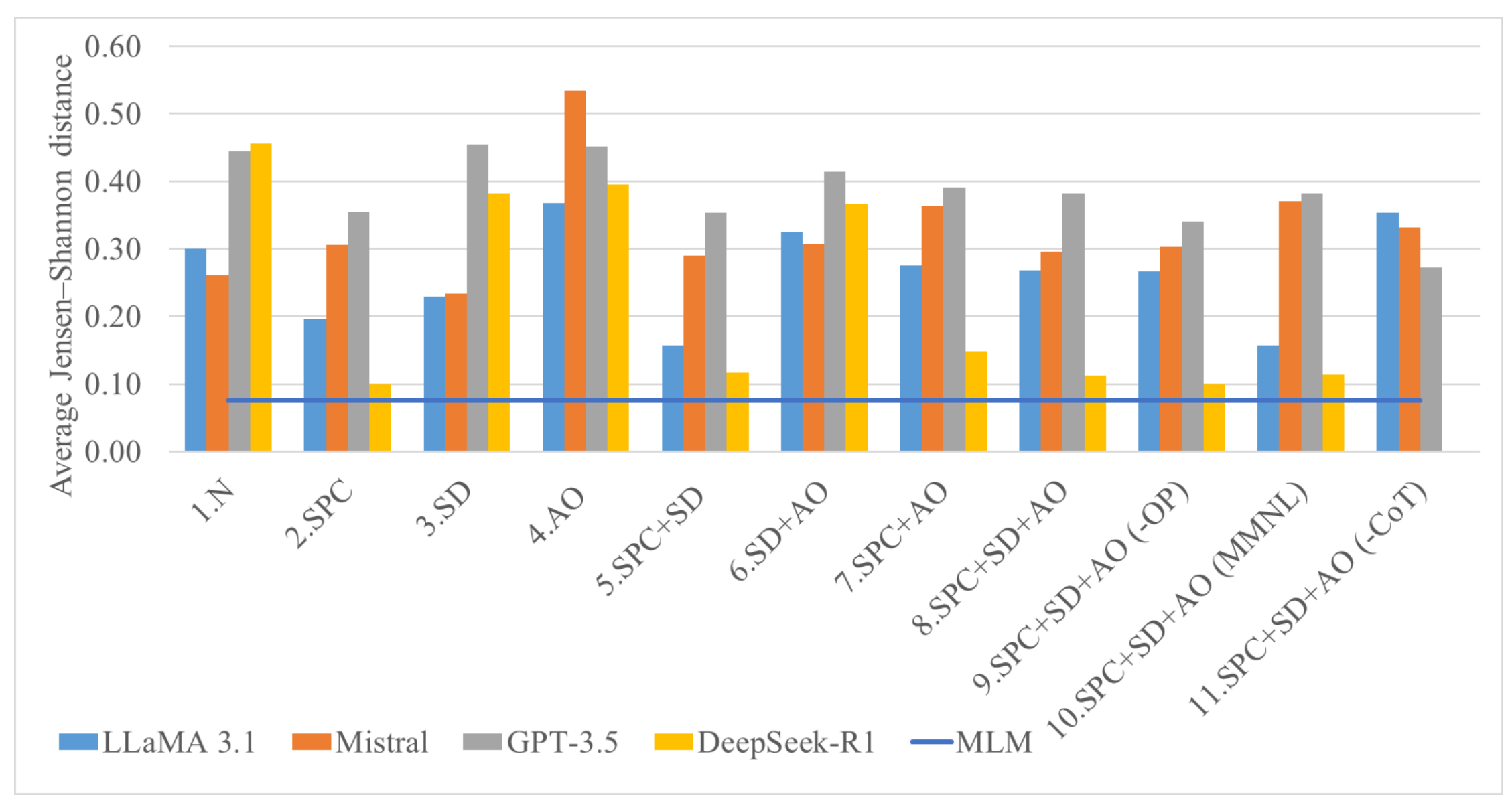

**Figure 4. Average Jensen–Shannon distance of simulated choice distribution**

### 4.4 Stratified analysis

This study stratified participants in the Heating SP Survey dataset based on several common household characteristics that are available, including (1) households with and without children under 18 years old, (2) households with and without older adults (65+ years old), and (3) households with annual income above or below £25,000. Since socio-demographic data were collected in Wave 1 of the Heating survey, which focused on household-level attributes, individual-level factors such as gender and age are not available. Appendix B.4 summarises the simulation accuracy of all LLMs within these stratified groups in Scenario 8, which was selected because all factors are present in its prompt. Table 7 presents the average accuracy and F1 score across three SP experiments and Figure 5 illustrates the variation in average accuracy with 95% confidence intervals.

Overall, the average accuracy across three SP choice experiments shows no substantial variation among the stratified groups within Scenario 8, indicating generally stable performance for the LLMs. However, notable differences arise when examining individual SP questions (see Appendix B.4). LLaMA 3.1 demonstrates lower accuracy in SP2 for the "With-Seniors" group while GPT-3.5 achieves higher accuracy in SP3 for the same group. In SP2, DeepSeek-R1 exhibits lower accuracy for the "With-Children" group but higher accuracy for "With- Seniors" groups. Such performance variability may reflect different representation of demographic groups in the LLM's training dataset.

**Table 7. Stratified analysis**

| Scenario/stratified groups | | LLaMA 3.1 | | Mistral | | GPT-3.5 | | DS-R1 | |
|---|---|---|---|---|---|---|---|---|---|
| | | Acc | F1 | Acc | F1 | Acc | F1 | Acc | F1 |
| 8 | SPC+SD+AO | 0.38 | 0.25 | 0.44 | 0.29 | 0.47 | 0.37 | 0.72 | 0.69 |
| | Child-Free [Size: 369] | 0.38 | 0.25 | 0.44 | 0.29 | 0.47 | 0.36 | 0.73 | 0.70 |
| | With-Children [Size: 192] | 0.38 | 0.25 | 0.43 | 0.29 | 0.45 | 0.37 | 0.70 | 0.68 |
| | No-Seniors [Size: 385] | 0.39 | 0.26 | 0.42 | 0.28 | 0.46 | 0.36 | 0.71 | 0.67 |

| | With-Seniors [Size: 176] | 0.37 | 0.24 | 0.46 | 0.34 | 0.48 | 0.38 | 0.76 | 0.73 |
|---|---|---|---|---|---|---|---|---|---|
| | Low-income [Size: 103] | 0.39 | 0.26 | 0.43 | 0.31 | 0.45 | 0.34 | 0.74 | 0.72 |
| | High-income [Size: 388] | 0.38 | 0.25 | 0.43 | 0.29 | 0.47 | 0.37 | 0.71 | 0.67 |

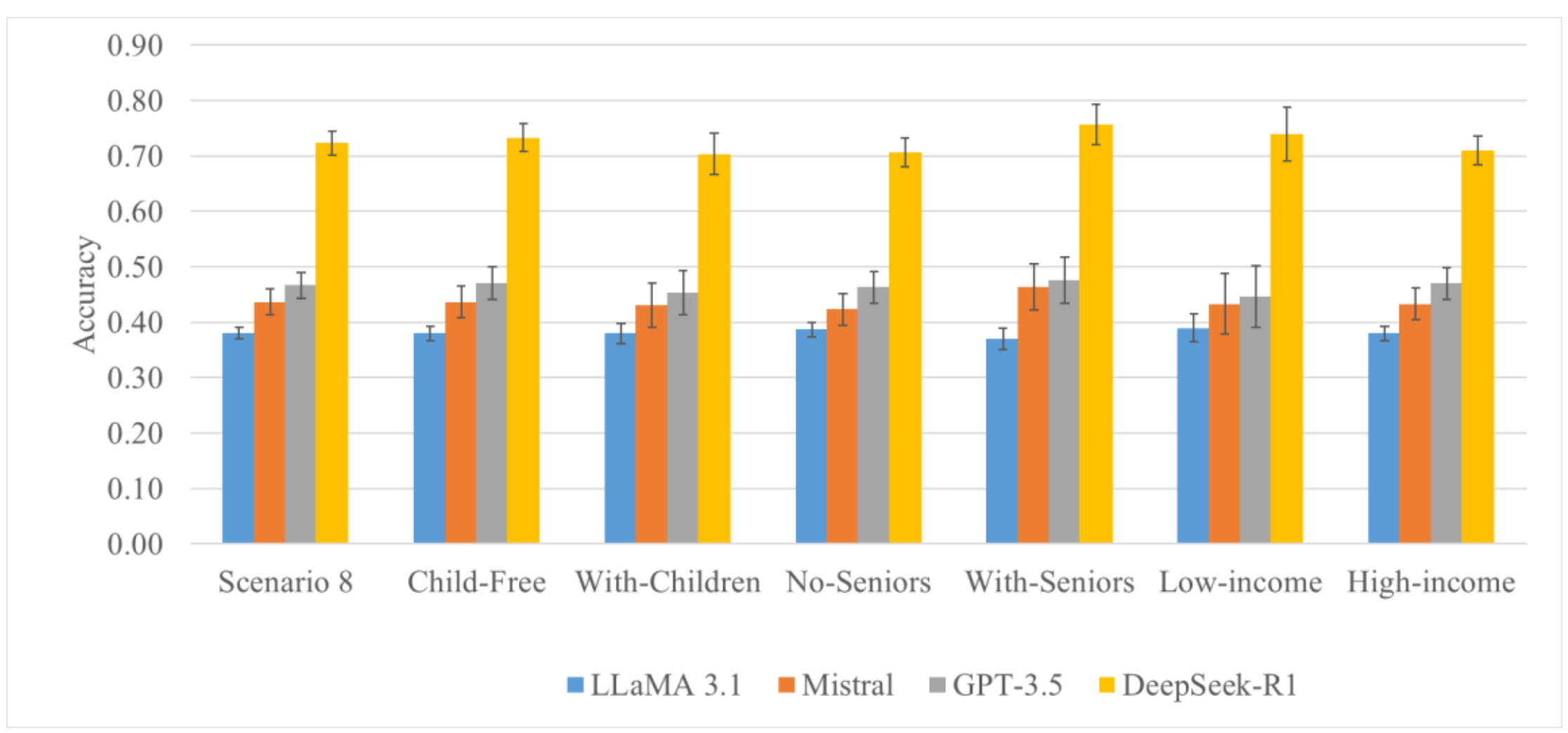

**Figure 5. Accuracy in stratified groups**

## 6. Discussion and conclusion

This study is among the first to explore the application of LLMs in simulating consumer choices in energy-related SP surveys. A series of test scenarios were designed to systematically evaluate the simulation performance of LLMs at both individual and aggregated levels, considering contexts such as factors in prompt design, ICL, CoT reasoning, the comparison between LLMs, integration with traditional choice models and potential sources of bias. Beyond prediction, this study also examined the potential of LLMs for contributing to both data collection and data analysis. The main findings, along with answers to the specific research questions, are as follows:

**Q1: To what extent can LLMs accurately simulate individuals' choices in an energy SP survey?**

1. The case study and test scenarios demonstrate that LLMs can simulate human response in energy-related SP surveys. DeepSeek-R1 achieves the highest average accuracy at 77%.

While the performance of all LLMs exceeds random guessing, it is sufficient only for specific applications, such as generating realistic responses, but not serving as a substitute for surveys. Its relatively strong performance without fine-tuning suggests potential for improvement through prompt optimisation, domain-specific fine-tuning, and the use of more advanced LLMs.

**Q2: How do different types of LLMs perform in energy SP survey simulations?**

2. Cloud-based LLMs do not consistently outperform local LLMs with smaller parameter sizes. In this case study, the reasoning LLM, DeepSeek-R1, significantly outperformed non-reasoning LLMs in terms of simulation accuracy, identification of insignificant factors, and alignment with observed choice distributions.

Local and cloud-based non-reasoning models (e.g., LLaMA 3.1, Mistral, and GPT-3.5) achieved comparable accuracy levels but frequently ignored valuable information in longer prompts and exhibited biased choice distributions. LLMs showed a systematic bias against less energy-efficient options (e.g., gas boiler and no retrofit). This may be linked to the intentional fine-tuning of commercial cloud-based models to align with socially desirable responses. Additionally, cloud-based LLMs (GPT-3.5 and DeepSeek-R1) showed greater consistency in following output format requirements, while smaller local models occasionally failed to meet the expected output structure. While DeepSeek-R1 demonstrates the best overall performance, it requires substantially longer processing times, approximately five times longer, compared to other LLMs.

Overall, these results highlight the advantages of reasoning LLMs for simulating choice behaviours and understanding consumer preferences more accurately.

**Q3: What is the optimal combination of factors to include in prompts for LLM simulations, considering individuals' previous choices, socio-demographic characteristics, attitudes and opinions?**

3.1 Individuals' previous SP choices are the most effective factor in prompts for LLM-based simulation in this case study.

3.2 Longer prompts with additional factors and varied formats can cause LLMs to lose focus, reducing accuracy.

Socio-demographics, attitudes and opinions do not significantly influence the simulation accuracy. The stratified analysis shows no strong demographic bias, except for some variation observed in the "With-Seniors" and "With-Children" groups for LLaMA3.1, GPT-3.5 and DeepSeek-R1. In addition, Scenario 9 confirms that LLMs already possess knowledge of common energy concepts from pretraining. Thus, explanations of standard terminology may be omitted after validation to simplify prompts.

**Q4: How can LLMs be integrated with traditional choice models and support data analysis in research?**

4.1 The traditional mixed logit choice model can help refine LLM prompts to improve simulation accuracy and choice distribution alignments.

4.2 LLMs, especially the reasoning LLM, can contribute to data analysis by indicating the significance of factors in predictive models, offering a qualitative complement to statistical models.

The mixed logit model, although not designed for simulation accuracy, achieves higher accuracy and aligns more closely with real consumer responses at the aggregate level than non-reasoning LLMs. It provides statistical indicators of factor influence, while LLMs offer flexible text-based reasoning. Importantly, mixed logit models require large and structured training datasets, while pre-trained LLMs can operate effectively with minimal task-specific data through prompt engineering. Despite these differences, LLMs can utilise findings from mixed logit models to refine prompts for specific tasks.

Notably, the reasoning LLM, DeepSeek-R1 exhibits the potential to infer the significance of factors through the ignored factor outputs in this study, offering a function analogous to significance testing in traditional choice modelling. This suggests LLMs could augment data analysis by providing both predictive capability and interpretable insights. Future research could explore LLMs' broader data analysis functions, such as generating statistical models by synthesising small relevant datasets with large-scale online information and public databases.

In summary, while LLMs fall short in accuracy and may introduce biases similar to surveys on real human respondents, they offer notable advantages in terms of survey efficiency, cost reduction, and scalability, all useful for applications such as rapid prototyping and simulations. Notably, reasoning LLMs show potential to support both data collection and data analysis, though further validation is required. Future research should analyse the explanatory components generated by LLMs to assess consistency and reasoning validity. Ideally, such studies should follow similar, comparable evaluation protocols. The format of factors within prompts requires deeper exploration, as the ignored factor analysis suggests potential conflicts when combining different input structures. Beyond ignored factor lists, alternative prompt designs or interpretability methods should be explored to more effectively extract reasoning and identify influential factors. These advancements will facilitate the integration of LLMs into general research workflows. Additionally, the effectiveness of CoT reasoning warrants further investigation. Although its broader impact is still under investigation [17,33], DeepSeek-R1 has demonstrated strong performance in this study, highlighting the need to further examine CoT's role in simulating human decision-making. Lastly, fine-tuning techniques should be examined to enhance accuracy and generalisability in energy-related applications.

**Declaration of Competing Interest**

The authors declare that they have no known competing financial interests or personal relationships that could have appeared to influence the work reported in this paper.

**Acknowledgements**

This work was supported by the Engineering and Physical Sciences Research Council (EPSRC) and the Economic and Social Research Council (ESRC) through funding provided to the Energy Demand Research Centre Project (grant number EP/Y010078/1).

**References:**

[1] Rekker L, Hulshof D, Kesina M, Mulder M. Willingness-to-pay for low-carbon residential heating systems: A discrete choice experiment among Dutch households. Energy and Buildings 2024;323:114712. https://doi.org/10.1016/j.enbuild.2024.114712.

[2] Naghiyev E, Shipman R, Goulden M, Gillott M, Spence A. Cost, context, or convenience? Exploring the social acceptance of demand response in the United Kingdom. Energy Research & Social Science 2022;87:102469. https://doi.org/10.1016/j.erss.2021.102469.

[3] Louviere JJ, Hensher DA, Swait JD, Adamowicz W. Stated Choice Methods: Analysis and Applications. 1st ed. Cambridge University Press; 2000. https://doi.org/10.1017/CBO9780511753831.

[4] Jansen BJ, Jung S, Salminen J. Employing large language models in survey research. Natural Language Processing Journal 2023;4:100020. https://doi.org/10.1016/j.nlp.2023.100020.
[5] Brand J, Israeli A, Ngwe D. Using LLMs for Market Research 2024.
[6] Zhang Y, Li Y, Cui L, Cai D, Liu L, Fu T, et al. Siren's Song in the AI Ocean: A Survey on Hallucination in Large Language Models 2023.
[7] Wei J, Wang X, Schuurmans D, Bosma M, Ichter B, Xia F, et al. Chain-of-Thought Prompting Elicits Reasoning in Large Language Models 2022. https://doi.org/10.48550/ARXIV.2201.11903.
[8] NESO. Clean Power 2030. NESO; 2024.
[9] Catapult. Electrification of Heat - Home Surveys and Install Report. Energy Systems Catapult 2020. https://es.catapult.org.uk/report/electrification-of-heat-home-surveys-and-install-report/ (accessed November 25, 2024).
[10] Ofgem. Consumer Survey 2021 | Ofgem 2023. https://www.ofgem.gov.uk/publications/consumer-survey-2021 (accessed November 25, 2024).
[11] Ofgem. Energy Consumer Satisfaction Survey: January to February 2024 | Ofgem 2024. https://www.ofgem.gov.uk/publications/energy-consumer-satisfaction-survey-january-february-2024 (accessed November 25, 2024).
[12] Centre for Sustainable Energy. How consumer archetypes shape the energy landscape. Centre for Sustainable Energy 2024. https://www.cse.org.uk/news/how-consumer-archetypes-shape-the-energy-landscape/ (accessed November 25, 2024).
[13] Ofgem. Ofgem Consumer Survey 2021 - Engagement report. 2021.
[14] Liu X, Yang D, Arentze T, Wielders T. The willingness of social housing tenants to participate in natural gas-free heating systems project: Insights from a stated choice experiment in the Netherlands. Applied Energy 2023;350:121706. https://doi.org/10.1016/j.apenergy.2023.121706.
[15] NERA Economic Consulting, Explain Market Research. Estimating Electricity and Gas Transmission Consumers' Willingness to Pay for Changes in Service during RIIO2. 2019.
[16] Lopez-Becerra EI, Alcon F. Social desirability bias in the environmental economic valuation: An inferred valuation approach. Ecological Economics 2021;184:106988. https://doi.org/10.1016/j.ecolecon.2021.106988.
[17] Wang M, Zhang DJ, Zhang H. Large Language Models for Market Research: A Data-augmentation Approach 2024. https://doi.org/10.48550/ARXIV.2412.19363.
[18] Almheiri SMAA, AlAnsari M, AlHashmi J, Abdalmajeed N, Jalil M, Ertek G. Data Analytics with Large Language Models (LLM): A Novel Prompting Framework. In: Emrouznejad A, Zervopoulos PD, Ozturk I, Jamali D, Rice J, editors. Business Analytics and Decision Making in Practice, Cham: Springer Nature Switzerland; 2024, p. 243–55. https://doi.org/10.1007/978-3-031-61589-4_20.
[19] Aher G, Arriaga RI, Kalai AT. Using Large Language Models to Simulate Multiple Humans and Replicate Human Subject Studies 2023.
[20] Webb T, Holyoak KJ, Lu H. Emergent analogical reasoning in large language models. Nat Hum Behav 2023;7:1526–41. https://doi.org/10.1038/s41562-023-01659-w.

[21] Strachan JWA, Albergo D, Borghini G, Pansardi O, Scaliti E, Gupta S, et al. Testing theory of mind in large language models and humans. Nat Hum Behav 2024;8:1285–95. https://doi.org/10.1038/s41562-024-01882-z.
[22] Luo X, Rechardt A, Sun G, Nejad KK, Yáñez F, Yilmaz B, et al. Large language models surpass human experts in predicting neuroscience results. Nat Hum Behav 2024;9:305–15. https://doi.org/10.1038/s41562-024-02046-9.
[23] Mancoridis M, Weeks B, Vafa K, Mullainathan S. Potemkin Understanding in Large Language Models 2025. https://doi.org/10.48550/arXiv.2506.21521.
[24] Goli A, Singh A. Frontiers: Can Large Language Models Capture Human Preferences? Marketing Science 2024;43:709–22. https://doi.org/10.1287/mksc.2023.0306.
[25] Argyle LP, Busby EC, Fulda N, Gubler J, Rytting C, Wingate D. Out of One, Many: Using Language Models to Simulate Human Samples. Polit Anal 2023;31:337–51. https://doi.org/10.1017/pan.2023.2.
[26] Qu Y, Wang J. Performance and biases of Large Language Models in public opinion simulation. Humanit Soc Sci Commun 2024;11:1095. https://doi.org/10.1057/s41599-024-03609-x.
[27] Fell M. Energy Social Surveys Replicated with Large Language Model Agents. SSRN Journal 2024. https://doi.org/10.2139/ssrn.4686345.
[28] Doshi-Velez F, Kim B. Towards A Rigorous Science of Interpretable Machine Learning 2017.
[29] Du M, Liu N, Hu X. Techniques for Interpretable Machine Learning 2019.
[30] Luo H, Specia L. From Understanding to Utilization: A Survey on Explainability for Large Language Models 2024.
[31] Zhao H, Chen H, Yang F, Liu N, Deng H, Cai H, et al. Explainability for Large Language Models: A Survey. ACM Trans Intell Syst Technol 2024;15:1–38. https://doi.org/10.1145/3639372.
[32] Liu F, Xu P, Li Z, Feng Y, Song H. Towards Understanding In-Context Learning with Contrastive Demonstrations and Saliency Maps 2024.
[33] Wu S, Shen EM, Badrinath C, Ma J, Lakkaraju H. Analyzing Chain-of-Thought Prompting in Large Language Models via Gradient-based Feature Attributions 2023.
[34] Jacek Pawlak, Han Wang, Aruna Sivakumar, Andreas Olympios, Matthias Mersch. UK Atittudes Towards Low-Carbon Domestic Heating Survey 2023-24 2024. https://doi.org/10.5281/zenodo.13626372.
[35] Jacek Pawlak, Han Wang, Andreas V. Olympios, Matthias Mersch, Aruna Sivakumar, Arthur Bessis, et al. Understanding preferences for adoption of low-carbon technologies for domestic heating in the UK unpublished results.
[36] Bouman T, Steg L, Kiers HAL. Measuring Values in Environmental Research: A Test of an Environmental Portrait Value Questionnaire. Front Psychol 2018;9:564. https://doi.org/10.3389/fpsyg.2018.00564.
[37] Li P-H, Keppo I, Strachan N. Incorporating homeowners' preferences of heating technologies in the UK TIMES model. Energy 2018;148:716–27. https://doi.org/10.1016/j.energy.2018.01.150.
[38] Li X, Yao R. Modelling heating and cooling energy demand for building stock using a hybrid approach. Energy and Buildings 2021;235:110740. https://doi.org/10.1016/j.enbuild.2021.110740.
[39] McFADDEN D, Train K. Mixed MNL Models for Discrete Response. Journal of Applied Econometrics 2000;15:447–70. https://www.jstor.org/stable/2678603.

[40] Train K. Discrete choice methods with simulation. 2nd ed. Cambridge ; New York: Cambridge University Press; 2009.
[41] Touvron H, Lavril T, Izacard G, Martinet X, Lachaux M-A, Lacroix T, et al. LLaMA: Open and Efficient Foundation Language Models 2023. https://doi.org/10.48550/ARXIV.2302.13971.
[42] Mündler N, He J, Jenko S, Vechev M. Self-contradictory Hallucinations of Large Language Models: Evaluation, Detection and Mitigation 2023. https://doi.org/10.48550/ARXIV.2305.15852.
[43] Salecha A, Ireland ME, Subrahmanya S, Sedoc J, Ungar LH, Eichstaedt JC. Large language models display human-like social desirability biases in Big Five personality surveys. PNAS Nexus 2024;3:pgae533. https://doi.org/10.1093/pnasnexus/pgae533.

## Appendix A: Example Prompts

The prompts in this study are automatically generated from the dataset using code. Given the large sample size and the diversity of individual factors, enforcing uniform academic writing conventions, whether manually or through automation, is not only challenging but also unnecessary. Despite their straightforward and unpolished structure, these prompts are fully interpretable by large language models and effectively elicit coherent responses.

A.1. Stated preference choice experiment: SP1

A.1.1 System message:

Imagine that you live in a large house built before 1965 with an area of approximately 110 square meters. You are now in a situation where you need to replace the current (natural) gas boiler, because it has reached its end-of-life or has malfunctioned, and repair is impossible. In this survey, you are presented with 3 distinct heating technologies to choose from: gas boiler, hydrogen ready boiler, and air source heat pump.

Brief descriptions of these technologies are available below:

Gas boiler: The traditional gas boiler utilises natural gas that is burnt to heat water, which is then circulated through radiators or underfloor heating systems to warm indoor spaces. Combustion of natural gas produces emissions, such as CO2, at the point of use. This technology has been around for a number of decades.

Hydrogen-ready boiler: A hydrogen-ready boiler is a gas-fired heating boiler which is capable of burning either natural gas or pure (100%) hydrogen. Hydrogen can be manufactured from water using electricity and is a carbon-free energy carrier. Combustion of hydrogen produces no carbon dioxide at the point of use. Furthermore, it is expected that the existing gas network can be adapted to supply hydrogen to properties. The technology is not yet available commercially but is expected to become available in the next few years.

Air source heat pump: An air source heat pump uses electricity to extract heat from the ambient air outside (even when it is cold, e.g. during winter) and transfers it indoors to heat water, which is then circulated to provide heating. The principles are very similar to how air conditioning works, though in reverse (to heat, rather than to cool). Heat pumps do not produce emissions at the point of use but consume electricity. The technology has been available for domestic heating for a number of decades.

These heating technology options will be described in terms of the following characteristics:

Fixed cost (in £): this is the one-off cost of the heating appliance itself, including installation.

Annual cost of operation (in £) and the equivalent average per month: this is how much you will typically need to pay in a year for the heating, including maintenance costs. (Financial)

Support scheme: this describes the availability of preferential loans (interest-free loan or 4% loan) to help with covering the cost of purchasing the specific heating option.

Maintenance visits: this describes how many visits by a skilled technician are needed each year to maintain the equipment. The cost is included in the annual cost of operation.

CO2 emissions (in kg per year): this describes how much carbon dioxide (CO2) is produced annually when using this heating technology, either directly (from combustion) or due to the generation of the required electricity. It is measured in kilograms of CO2, but we also present for comparison an equivalent i.e. CO2 emissions in terms of the number of single person economy flights from London to Glasgow.

Based on the above characteristics, we will be asking you to select one option that you find the most attractive in each scenario.

**Your previous choices**: The options, their characteristics and your choice based on your preference are:

[ *(Scenario 1 as an example)*

```
{
  "options": [
    {
      "name": "Gas boiler",
      "characteristics": {
        "Fixed cost": "£ 1900",
        "Annual cost of operation (in £) and the equivalent average per month": "£ 3600 (£ 300 per month)",
        "Support scheme": "None",
        "Maintenance visits": "1 per year",
        "CO2 emission (Equivalent number of single-person economy flights from London to Glasgow)": "4590 kg per year (26 flight(s))"
      }
    },
    {
      "name": "Hydrogen ready boiler",
      "characteristics": {
        "Fixed cost": "£ 1900",
        "Annual cost of operation (in £) and the equivalent average per month": "£ 6600 (£ 550 per month)",
        "Support scheme": "4% loan",
        "Maintenance visits": "2 per year",
        "CO2 emission (Equivalent number of single-person economy flights from London to Glasgow)": "1000 kg per year (6 flight(s))"
      }
    },
    {
```

```
      "name": "Air source heat pump",
      "characteristics": {
        "Fixed cost": "£ 15300",
        "Annual cost of operation (in £) and the equivalent average per month": "£ 1680 (£ 140 per month)",
        "Support scheme": "Interest-free loan",
        "Maintenance visits": "2 per year",
        "CO2 emission (Equivalent number of single-person economy flights from London to Glasgow)": "0 kg per year (0 flight(s))"
      }
    }
  ],
  "Choice": "Hydrogen ready boiler"
}
```

}, {… *(Scenario2)* …},{…*(Scenario3)* … },{…*(Scenario4)* … },{…*(Scenario5)* … }]

**Your socio-demographics**:

The overall combined annual income of all people living in your property is £10,000-14,999. Your household members include 1 person aged 0-5 years old, 1 person aged 6-12 years old, 1 person aged 13-17 years old, 1 person aged 18-65 years old, and 0 person aged more than 65 years old. 1 individuals living in the property are employed. The highest level of education among all individuals living in your property is a Secondary school. Your property is Semi-detached, with 8 rooms, 101-150 m² total usable floor area, with LED lights, built Between 1945 and 1964, located in Yorkshire and the Humber, with an EPC rating of B. You have spent £101-200 on inspections and services over the past two years. The primary source of heating in your property is a Gas boiler, which is installed 4-6 years ago. You have an electricity tariff: Fixed price (Price of energy is fixed for 12 or 24 months).

**Your personality and attitudes**:

You find a person for whom it is important to prevent environmental pollution, Mostly like you. You find a person for whom it is important to protect the environment, Mostly like you. You find a person for whom it is important to respect nature, Very much like you. You find a person for whom it is important to be in unity with nature, Mostly like you. You find a person for whom it is important to have control over others' actions, Very much like you. You find a person for whom it is important to have authority over others, Very much like you. You find a person for whom it is important to be influential, Very much like you. You find a person for whom it is important to have money and possessions, Mostly like you. You find a person for whom it is important to work hard and be ambitious, Very much like you. You find a person for whom it is important to have fun, Mostly like you. You find a person for whom it is important to enjoy life's pleasures, Mostly like you. You find a person for whom it is important to do things he/she enjoys, Mostly like you. You find a person for whom it is important that everybody has equal opportunities, Very much like you. You find a person for whom it is important to take care of those who are worse off, Mostly like you. You find a person for whom it is important that everybody is treated justly, Very much like you. You find a person for whom it is important that there is no war or conflict, Very much like you. You find a person for whom it is important to be helpful to others, Very much like you. You Somewhat agree that you can obtain reliable advice or guidance about greener and more energy-efficient heating options. You Somewhat agree that you are confident you would get a high-quality installation. You Strongly agree that you are clear about the expected level of disruption. You Strongly agree that the up-front costs of appliances and the installation will be affordable. You Somewhat agree that the bills and maintenance of the heating system will be affordable. You Somewhat agree that the more environmentally friendly heating system would meet my heating needs. You Most of the time check the energy rating on appliances before purchase. You Always reduced waste and increased recycling. You Most of the time adapt your food shopping habits to buy less carbon intensive food. You Always used active travel and public transport alternatives for most travel. You Always taken holidays in the UK rather than abroad specifically to avoid air travel.

### A.1.2 User message

Now the options and their characteristics are provided below:

```
{
  "options": [
    {
      "name": "Gas boiler",
      "characteristics": {
        "Fixed cost": "£ 1900",
        "Annual cost of operation (in £) and the equivalent average per month":
"£ 6240 (£ 520 per month)",
        "Support scheme": "4% loan",
        "Maintenance visits": "2 per year",
        "CO2 emission (Equivalent number of single-person economy flights from London
to Glasgow)": "4080 kg per year (23 flight(s))"
      }
    },
    {
      "name": "Hydrogen ready boiler",
      "characteristics": {
        "Fixed cost": "£ 1900",
        "Annual cost of operation (in £) and the equivalent average per month":
"£ 1920 (£ 160 per month)",
        "Support scheme": "None",
        "Maintenance visits": "2 per year",
        "CO2 emission (Equivalent number of single-person economy flights from London
to Glasgow)": "0 kg per year (0 flight(s))"
      }
```

```
    },
    {
      "name": "Air source heat pump",
      "characteristics": {
        "Fixed cost": "£ 15300",
        "Annual cost of operation (in £) and the equivalent average per month": "£ 2880 (£ 240 per month)",
        "Support scheme": "Interest-free loan",
        "Maintenance visits": "1 per year",
        "CO2 emission (Equivalent number of single-person economy flights from London to Glasgow)": "2020 kg per year (12 flight(s))"
      }
    }
  ],
  "Choice": null
}
```

Please select one of the options based on these provided factors: [' Your previous choices', ' Your socio-demographics', ' Your personality and attitudes']?

Respond in valid JSON format only. The first JSON object, called "Explanation", is a short (<50 words) explanation of your reasoning of choice, based on your preference. The second JSON object, called "Choice", shows your choice between three options. Use response codes: gas boiler=1 or hydrogen ready boiler = 2 or air source heat pump=3. Your decision must be consistent with your explanation. The third JSON object, called "Ignored", is in a list format and shows the name of which part of the system message and user message you don't consider in making the choice. Example output as follows: {"Explanation": "Explanation text here", "Choice": 1 or 2 or 3, "Ignored":["part1", "part2"]}

A.2. Stated preference choice experiment: SP2

A.2.1 System message:

This part of the survey looks at retrofitting options for your property, which can accompany the replacement of the heating technology. By retrofit, we mean additional upgrades to the property that can help in making it more energy efficient, i.e. consume less energy whilst maintaining the same comfort. The context involves you living in a large house built before 1965 with an area of approximately 110 square meters. In this part of the survey, we will focus on 3 options for retrofit that can accompany the heating appliance installation: no retrofit, minor retrofit and major retrofit. These options are described below. In the current context, the retrofit options will accompany any heating technology of your choice and the heating technology is therefore not specified in the following scenarios.

No retrofit: You only replace the heating appliance.

Minor retrofit: In addition to replacing your heating appliance, you decide to undertake some minor upgrades such as draught proofing, reduced infiltration or hot water tank insulation. In exchange, you reduce your energy consumption.

Major retrofit: In addition to replacing your heating appliance and minor retrofit upgrades, you also decide to improve wall, floor and loft insulation as well as improved window insulation (double or triple glazing). This is all more expensive and causes more nuisance but leads to more substantial energy savings.

These retrofit options will be described in terms of the following characteristics:

Equipment cost (in £): this is the one-off cost of the heating appliance itself, including installation; and will remain fixed for each scenario.

Retrofit cost (in £): this is the one-off cost of the retrofit, including parts and labour.

Nuisance duration (in days): this is how many days you may expect installation or retrofit works to be taking place. This may involve contractors working in your property and carrying out the retrofit works.

Savings on monthly costs of heating (in £) and payback period (in years): this is how much you can expect to save on your energy bills as a result of the retrofit and the approximate period over which the retrofit cost will be recovered by these savings.

(Financial) Support scheme: this describes the availability of preferential loans (interest-free loan or 4% loan) to help with covering the cost of the heating technology replacement and retrofit works.

Based on the above characteristics, we will be asking you to select one option that you find the most attractive in each scenario.

Your previous choices: The options, their characteristics and your choice based on your preference are below:

[ *(Scenario 1 as an example)*

```
{
  "options": [
    {
      "name": "No retrofit",
      "characteristics": {
        "Equipment cost": "£ 10400",
        "Retrofit cost": null,
        "Nuisance duration": "1 day(s)",
        "Savings on monthly costs of heating & payback period": null,
        "Support scheme": "None"
      }
    },
    {
```

```
      "name": "Minor retrofit",
      "characteristics": {
        "Equipment cost": "£ 10400",
        "Retrofit cost": "£ 430",
        "Nuisance duration": "1 day(s)",
        "Savings on monthly costs of heating & payback period": "£ 25 (Over 1 year(s))",
        "Support scheme": "None"
      }
    },
    {
      "name": "Major retrofit",
      "characteristics": {
        "Equipment cost": "£ 10400",
        "Retrofit cost": "£ 12000",
        "Nuisance duration": "30 day(s)",
        "Savings on monthly costs of heating & payback period": "£ 116 (Over 9 year(s))",
        "Support scheme": "None"
      }
    }
  ],
  "Choice": "Minor retrofit"
}
```

}, {… *(Scenario2)* …},{…*(Scenario3)* … },{…*(Scenario4)* … },{…*(Scenario5)* … }]

**Your socio-demographics**:

The overall combined annual income of all people living in your property is £10,000-14,999. Your household members include 1 person aged 0-5 years old, 1 person aged 6-12 years old, 1 person aged 13-17 years old, 1 person aged 18-65 years old, and 0 person aged more than 65 years old. 1 individuals living in the property are employed. The highest level of education among all individuals living in your property is a Secondary school. Your property is Semi-detached, with 8 rooms, 101-150 m² total usable floor area, with LED lights, built Between 1945 and 1964, located in Yorkshire and the Humber, with an EPC rating of B. You have spent £101-200 on inspections and services over the past two years. The primary source of heating in your property is a Gas boiler, which is installed 4-6 years ago. You have an electricity tariff: Fixed price (Price of energy is fixed for 12 or 24 months).

**Your personality and attitudes**:

You find a person for whom it is important to prevent environmental pollution, Mostly like you. You find a person for whom it is important to protect the environment, Mostly like you. You find a person for whom it is important to respect nature, Very much like you. You find a person for whom it is important to be in unity with nature, Mostly like you. You find a person for whom it is important to have control over others' actions, Very much like you. You find a person for whom it is important to have authority over others, Very much like you. You find a person for whom it is important to be influential, Very much like you. You find a person for whom it is important to have money and possessions, Mostly like you. You find a person for whom it is important to work hard and be ambitious, Very much like you. You find a person for whom it is important to have fun, Mostly like you. You find a person for whom it is important to enjoy life's pleasures, Mostly like you. You find a person for whom it is important to do things he/she enjoys, Mostly like you. You find a person for whom it is important that everybody has equal opportunities, Very much like you. You find a person for whom it is important to take care of those who are worse off, Mostly like you. You find a person for whom it is important that everybody is treated justly, Very much like you. You find a person for whom it is important that there is no war or conflict, Very much like you. You find a person for whom it is important to be helpful to others, Very much like you. You Somewhat agree that you can obtain reliable advice or guidance about greener and more energy-efficient heating options. You Somewhat agree that you are confident you would get a high-quality installation. You Strongly agree that you are clear about the expected level of disruption. You Strongly agree that the up-front costs of appliances and the installation will be affordable. You Somewhat agree that the bills and maintenance of the heating system will be affordable. You Somewhat agree that the more environmentally friendly heating system would meet my heating needs. You Most of the time check the energy rating on appliances before purchase. You Always reduced waste and increased recycling. You Most of the time adapt your food shopping habits to buy less carbon intensive food. You Always used active travel and public transport alternatives for most travel. You Always taken holidays in the UK rather than abroad specifically to avoid air travel.

A.2.2 User message:

Now the options and their characteristics are provided below:

```
{
  "options": [
    {
      "name": "No retrofit",
      "characteristics": {
        "Equipment cost": "£ 10400",
        "Retrofit cost": null,
        "Nuisance duration": "1 day(s)",
        "Savings on monthly costs of heating & payback period": null,
        "Support scheme": "Interest-free loan"
      }
    },
```

```
    {
      "name": "Minor retrofit",
      "characteristics": {
        "Equipment cost": "£ 10400",
        "Retrofit cost": "£ 250",
        "Nuisance duration": "1 day(s)",
        "Savings on monthly costs of heating & payback period": "£ 30 (Over 1 year(s))",
        "Support scheme": "Interest-free loan"
      }
    },
    {
      "name": "Major retrofit",
      "characteristics": {
        "Equipment cost": "£ 10400",
        "Retrofit cost": "£ 41000",
        "Nuisance duration": "30 day(s)",
        "Savings on monthly costs of heating & payback period": "£ 96 (Over 36 year(s))",
        "Support scheme": "Interest-free loan"
      }
    }
  ],
  "Choice": null
}
```

Please select one of the options based on these provided factors: [' Your previous choices', ' Your socio-demographics', ' Your personality and attitudes']?

Respond in valid JSON format only. The first JSON object, called "Explanation", is a short (<50 words) explanation of your reasoning of choice, based on your preference. The second JSON object, called "Choice", shows your choice between three options. Use response codes: No retrofit=1 or Minor retrofit = 2 or Major retrofit=3. Your decision must be consistent with your explanation. The third JSON object, called "Ignored", is in a list format and shows the name of which part of the system message and user message you don't consider in making the choice. Example output as follows: {"Explanation": "Explanation text here", "Choice": 1 or 2 or 3, "Ignored":["part1", "part2"]}

A.3. Stated preference choice experiment: SP3

A.3.1 System message:

This third part of the survey focuses on the ownership model for the heating system, i.e. who owns the appliances heating your property and how the system is controlled and paid for. The context involves you living in a large house built before 1965 with an area of approximately 110 square meters. In this part of the survey, we will focus on 3 ownership options for the heating appliance: full ownership, shared ownership and service-based. These options are described below. In the current context, the ownership options are not specific to the type of heating technology and apply to any heating technology that you prefer.

Full ownership: You fully own the heating system and you are responsible for the maintenance of the equipment and its operation.

Shared ownership: You share ownership of a larger heating system with your neighbours/community. This reduces the overall cost of equipment and operation, but you potentially have less control over how the system is operated and maintained.

Service-based: You do not need to purchase the equipment yourself as this is done by the service provider. Since you do not own the equipment there is no upfront cost. In exchange, you may need to pay more on a monthly basis, for the duration of the contract that you sign with the service provider.

These ownership models will be described in terms of the following characteristics:

Upfront cost (in £): this is the one-off cost that you need to invest upfront, including equipment and installation.

Annual cost of operation (in £): this is how much you will need to typically pay for the heating in a year (and the equivalent per month cost).

Contract length (in years): this attribute tells you how many years you are contractually bound to a specific heating service provider and service.

Energy pricing: this tells you whether the price of energy used to heat your home is fixed or may vary during the day, and how much in advance you are informed about the energy prices. The following arrangements are possible: (1) fixed, i.e. the prices are constant across hours and days. (2) can change during the day and you are informed 24 hours ahead of any changes. (3) can change during the day and you are informed 1 hour ahead of any changes.

Control and service flexibility: this describes whether the operator can switch off the heating for 1 hour once a month, always informing you ahead of time, and how much you would be compensated for that. The following arrangements are possible: (1) You control the heating entirely, with no switching-off events. (2) Heating can be switched off for 1 hour a month, with notification 7 hours ahead. The compensation in pounds (£) per such event will be shown. (3) Heating can be switched off for 1 hour a month, with notification 1 hour ahead. The compensation in pounds (£) per such event will be shown. Based on the above characteristics, we will be asking you to select one option that you find the most attractive in each scenario.

Your previous choices: The options, their characteristics and your choice based on your preference are below:

[ *(Scenario 1 as an example)*

```
{
  "options": [
```

```
      {
        "name": "Full ownership",
        "characteristics": {
          "Upfront cost": "£ 10400",
          "Annual cost of operation (Average per month)": "£ 3600 (£ 300 per month)",
          "Contract length": null,
          "Energy pricing": "Can change during the day and you are informed 1 hour ahead of any changes",
          "Control and service flexibility": "Heating can be switched off for 1 hour a month. Notification 1 hour ahead. Compensation £14"
        }
      },
      {
        "name": "Shared ownership",
        "characteristics": {
          "Upfront cost": "£ 5700",
          "Annual cost of operation (Average per month)": "£ 3600 (£ 300 per month)",
          "Contract length": "4 year(s)",
          "Energy pricing": "Can change during the day and you are informed 24 hours ahead of any changes",
          "Control and service flexibility": "Heating can be switched off for 1 hour a month. Notification 7 hours ahead. Compensation £14"
        }
      },
      {
        "name": "Service-based",
        "characteristics": {
          "Upfront cost": null,
          "Annual cost of operation (Average per month)": "£ 4920 (£ 410 per month)",
          "Contract length": "1 year(s)",
          "Energy pricing": "Constant across hours and days",
          "Control and service flexibility": "You control the heating entirely"
        }
      }
    ],
    "Choice": "Shared ownership"
  }
```

}, {… *(Scenario2)* …},{…*(Scenario3)* … },{…*(Scenario4)* … },{…*(Scenario5)* … }]

**Your socio-demographics**:

The overall combined annual income of all people living in your property is £10,000-14,999. Your household members include 1 person aged 0-5 years old, 1 person aged 6-12 years old, 1 person aged 13-17 years old, 1 person aged 18-65 years old, and 0 person aged more than 65 years old. 1 individuals living in the property are employed. The highest level of education among all individuals living in your property is a Secondary school. Your property is Semi-detached, with 8 rooms, 101-150 m² total usable floor area, with LED lights, built Between 1945 and 1964, located in Yorkshire and the Humber, with an EPC rating of B. You have spent £101-200 on inspections and services over the past two years. The primary source of heating in your property is a Gas boiler, which was installed 4-6 years ago. You have an electricity tariff: Fixed price (Price of energy is fixed for 12 or 24 months).

**Your personality and attitudes**:

You find a person for whom it is important to prevent environmental pollution, Mostly like you. You find a person for whom it is important to protect the environment, Mostly like you. You find a person for whom it is important to respect nature, Very much like you. You find a person for whom it is important to be in unity with nature, Mostly like you. You find a person for whom it is important to have control over others' actions, Very much like you. You find a person for whom it is important to have authority over others, Very much like you. You find a person for whom it is important to be influential, Very much like you. You find a person for whom it is important to have money and possessions, Mostly like you. You find a person for whom it is important to work hard and be ambitious, Very much like you. You find a person for whom it is important to have fun, Mostly like you. You find a person for whom it is important to enjoy life's pleasures, Mostly like you. You find a person for whom it is important to do things he/she enjoys, Mostly like you. You find a person for whom it is important that everybody has equal opportunities, Very much like you. You find a person for whom it is important to take care of those who are worse off, Mostly like you. You find a person for whom it is important that everybody is treated justly, Very much like you. You find a person for whom it is important that there is no war or conflict, Very much like you. You find a person for whom it is important to be helpful to others, Very much like you. You Somewhat agree that you can obtain reliable advice or guidance about greener and more energy-efficient heating options. You Somewhat agree that you are confident you would get a high-quality installation. You Strongly agree that you are clear about the expected level of disruption. You Strongly agree that the up-front costs of appliances and the installation will be affordable. You Somewhat agree that the bills and maintenance of the heating system will be affordable. You Somewhat agree that the more environmentally friendly heating system would meet my heating needs. You Most of the time check the energy rating on appliances before purchase. You Always reduced waste and increased recycling. You Most of the time adapt your food shopping habits to buy less carbon intensive food. You Always used active travel and public transport alternatives for most travel. You Always taken holidays in the UK rather than abroad specifically to avoid air travel.

A.3.2 User message:

Now the options and their characteristics are provided below:

```
{
  "options": [
    {
      "name": "Full ownership",
      "characteristics": {
        "Upfront cost": "£ 10400",
        "Annual cost of operation (Average per month)": "£ 3600 (£ 300 per month)",
        "Contract length": null,
        "Energy pricing": "Can change during the day and you are informed 1 hour ahead of any changes",
        "Control and service flexibility": "You control the heating entirely"
      }
    },
    {
      "name": "Shared ownership",
      "characteristics": {
        "Upfront cost": "£ 5700",
        "Annual cost of operation (Average per month)": "£ 3600 (£ 300 per month)",
        "Contract length": "4 year(s)",
        "Energy pricing": "Constant across hours and days",
        "Control and service flexibility": "Heating can be switched off for 1 hour a month. Notification 1 hour ahead. Compensation £14"
      }
    },
    {
      "name": "Service-based",
      "characteristics": {
        "Upfront cost": null,
        "Annual cost of operation (Average per month)": "£ 4920 (£ 410 per month)",
        "Contract length": "1 year(s)",
        "Energy pricing": "Can change during the day and you are informed 24 hours ahead of any changes",
        "Control and service flexibility": "Heating can be switched off for 1 hour a month. Notification 7 hours ahead. Compensation £14"
      }
    }
  ],
  "Choice": null
}
```

Please select one of the options based on these provided factors: [' Your previous choices', ' Your socio-demographics', ' Your personality and attitudes']?

Respond in valid JSON format only. The first JSON object, called "Explanation", should be a short (<50 words) explanation of your reasoning of choice, based on your preference. The second JSON object, called "Choice", shows your choice between three options. Use response codes: Full ownership=1 or Shared ownership = 2 or Service-based=3. Your decision must be consistent with your explanation. The third JSON object, called "Ignored", is in a list format and shows the name of which part of the system message and user message you don't consider in making the choice. Example output as follows: {"Explanation": "Explanation text here", "Choice": 1 or 2 or 3, "Ignored":["part1", "part2"]}

## Appendix B: Detailed test results table

**Table B1.** Scenario ID and code

| Scenario ID | Components |
|---|---|
| 1 | N (default) |
| 2 | SPC |
| 3 | SD |
| 4 | AO |
| 5 | SPC+SD |
| 6 | SD+AO |
| 7 | SPC+AO |
| 8 | SPC+SD+AO |
| 9 | SPC+SD+AO (-OP) |
| 10 | SPC+SD+AO (MMNL) |
| 11 | SPC+SD+AO (-CoT) |

B.1. Individual prediction accuracy

**Table B2. Results of LLaMA 3.1***

| Scenario | SP1 | | SP2 | | SP3 | |
|---|---|---|---|---|---|---|
| | Acc | F1 | Acc | F1 | Acc | F1 |
| 1 | 0.45±0.0182 | 0.26 | 0.42±0.0182 | 0.20 | 0.38±0.018 | 0.25 |
| 2 | 0.44±0.0185 | 0.29 | 0.49±0.0185 | 0.29 | 0.51±0.0185 | 0.36 |
| 3 | 0.48±0.0184 | 0.32 | 0.40±0.018 | 0.21 | 0.35±0.0176 | 0.26 |
| 4 | 0.39±0.0178 | 0.22 | 0.22±0.0155 | 0.15 | 0.32±0.0173 | 0.23 |
| 5 | 0.42±0.0182 | 0.31 | 0.45±0.0184 | 0.23 | 0.39±0.018 | 0.29 |
| 6 | 0.40±0.0182 | 0.25 | 0.31±0.0173 | 0.18 | 0.31±0.0171 | 0.23 |
| 7 | 0.40±0.0178 | 0.24 | 0.37±0.0175 | 0.20 | 0.41±0.0182 | 0.28 |
| 8 | 0.39±0.018 | 0.26 | 0.36±0.0178 | 0.20 | 0.39±0.0182 | 0.28 |
| 9 | 0.40±0.0178 | 0.26 | 0.37±0.0182 | 0.22 | 0.40±0.0182 | 0.29 |
| 10 | 0.42±0.0182 | 0.28 | 0.54±0.0182 | 0.34 | 0.40±0.0182 | 0.29 |
| 11 | 0.38±0.0178 | 0.19 | 0.40±0.018 | 0.16 | 0.35±0.0176 | 0.21 |

*Results over 5 runs of LLama3.1

**Table B3. Results of Mistral**

| Scenario | SP1 | | SP2 | | SP3 | |
|---|---|---|---|---|---|---|
| | Acc | F1 | Acc | F1 | Acc | F1 |
| 1 | 0.45±0.041 | 0.30 | 0.42±0.041 | 0.18 | 0.35±0.0392 | 0.24 |
| 2 | 0.44±0.041 | 0.31 | 0.49±0.041 | 0.24 | 0.56±0.041 | 0.40 |
| 3 | 0.43±0.041 | 0.32 | 0.44±0.0419 | 0.20 | 0.33±0.0392 | 0.24 |
| 4 | 0.38±0.041 | 0.23 | 0.12±0.0267 | 0.08 | 0.32±0.0374 | 0.20 |
| 5 | 0.44±0.041 | 0.33 | 0.47±0.041 | 0.18 | 0.43±0.041 | 0.33 |
| 6 | 0.45±0.041 | 0.34 | 0.31±0.0383 | 0.17 | 0.31±0.0383 | 0.22 |
| 7 | 0.39±0.0401 | 0.25 | 0.46±0.041 | 0.25 | 0.40±0.0401 | 0.30 |
| 8 | 0.40±0.041 | 0.29 | 0.47±0.041 | 0.26 | 0.44±0.041 | 0.33 |
| 9 | 0.42±0.041 | 0.31 | 0.47±0.041 | 0.25 | 0.44±0.041 | 0.33 |
| 10 | 0.38±0.0401 | 0.26 | 0.48±0.041 | 0.25 | 0.41±0.041 | 0.31 |
| 11 | 0.43±0.0401 | 0.30 | 0.51±0.041 | 0.29 | 0.38±0.0392 | 0.26 |

**Table B4. Results of GPT-3.5**

| Scenario | SP1 | | SP2 | | SP3 | |
|---|---|---|---|---|---|---|
| | Acc | F1 | Acc | F1 | Acc | F1 |
| 1 | 0.36±0.0392 | 0.29 | 0.42±0.041 | 0.26 | 0.34±0.0392 | 0.27 |
| 2 | 0.40±0.041 | 0.32 | 0.51±0.041 | 0.35 | 0.60±0.0401 | 0.52 |
| 3 | 0.35±0.0392 | 0.28 | 0.47±0.041 | 0.25 | 0.34±0.0383 | 0.27 |
| 4 | 0.35±0.0392 | 0.27 | 0.20±0.0339 | 0.17 | 0.42±0.041 | 0.32 |
| 5 | 0.43±0.041 | 0.34 | 0.48±0.041 | 0.27 | 0.61±0.0401 | 0.56 |
| 6 | 0.35±0.0392 | 0.28 | 0.40±0.041 | 0.26 | 0.40±0.0401 | 0.30 |
| 7 | 0.37±0.0401 | 0.29 | 0.36±0.0392 | 0.27 | 0.57±0.041 | 0.50 |
| 8 | 0.38±0.0401 | 0.30 | 0.44±0.041 | 0.32 | 0.58±0.041 | 0.49 |
| 9 | 0.43±0.041 | 0.34 | 0.53±0.041 | 0.37 | 0.57±0.041 | 0.52 |
| 10 | 0.42±0.041 | 0.33 | 0.49±0.041 | 0.28 | 0.54±0.041 | 0.45 |
| 11 | 0.37±0.0401 | 0.30 | 0.46±0.041 | 0.31 | 0.30±0.0374 | 0.29 |

**Table B5. Results of DeepSeek-R1**

| Scenario | SP1 | | SP2 | | SP3 | |
|---|---|---|---|---|---|---|
| | Acc | F1 | Acc | F1 | Acc | F1 |
| 1 | 0.55±0.041 | 0.45 | 0.50±0.0419 | 0.23 | 0.35±0.0392 | 0.23 |
| 2 | 0.71±0.0374 | 0.71 | 0.84±0.0303 | 0.58 | 0.76±0.0348 | 0.74 |
| 3 | 0.55±0.041 | 0.50 | 0.51±0.041 | 0.25 | 0.34±0.0392 | 0.32 |
| 4 | 0.47±0.041 | 0.38 | 0.48±0.041 | 0.22 | 0.33±0.0383 | 0.31 |
| 5 | 0.72±0.0374 | 0.72 | 0.82±0.0321 | 0.75 | 0.71±0.0374 | 0.69 |
| 6 | 0.51±0.041 | 0.47 | 0.50±0.041 | 0.27 | 0.31±0.0383 | 0.30 |
| 7 | 0.59±0.0401 | 0.57 | 0.80±0.033 | 0.73 | 0.72±0.0365 | 0.70 |
| 8 | 0.64±0.0392 | 0.64 | 0.81±0.0321 | 0.73 | 0.72±0.0365 | 0.70 |
| 9 | 0.65±0.0392 | 0.65 | 0.80±0.033 | 0.71 | 0.72±0.0366 | 0.71 |
| 10 | 0.66±0.0392 | 0.66 | 0.81±0.0321 | 0.76 | 0.75±0.0357 | 0.73 |

B.2. Ignoreed factors

**Table B6. Percentage of simulated responses generated by non-reasoning LLMs containing ignored factors**

| | | LLaMA 3.1 | | | | | | | | |
|---|---|---|---|---|---|---|---|---|---|---|
| | | SP1 | | | SP2 | | | SP3 | | |
| | | SPC | AO | SD | SPC | AO | SD | SPC | AO | SD |
| 8 | SPC+SD+AO | 75% | 37% | 28% | 76% | 36% | 16% | 80% | 41% | 24% |
| 9 | SPC+SD+AO (-OP) | 75% | 37% | 28% | 76% | 38% | 17% | 77% | 43% | 24% |
| 10 | SPC+SD+AO(MMNL) | 84% | 42% | 17% | 31% | 56% | 48% | 81% | 64% | 19% |
| 11 | SPC+SD+AO(-CoT) | 82% | 12% | 17% | 73% | 17% | 9% | 80% | 17% | 11% |
| | | | | | | | | | | |
| | | Mistral | | | | | | | | |
| | | SP1 | | | SP2 | | | SP3 | | |
| | | SPC | AO | SD | SPC | AO | SD | SPC | AO | SD |
| 8 | SPC+SD+AO | 79% | 22% | 46% | 67% | 18% | 27% | 80% | 33% | 47% |
| 9 | SPC+SD+AO (-OP) | 80% | 24% | 46% | 71% | 19% | 28% | 78% | 37% | 47% |
| 10 | SPC+SD+AO(MMNL) | 73% | 25% | 50% | 19% | 20% | 32% | 70% | 42% | 50% |
| 11 | SPC+SD+AO(-CoT) | 67% | 18% | 32% | 66% | 15% | 21% | 62% | 19% | 35% |
| | | | | | | | | | | |
| | | GPT-3.5 | | | | | | | | |
| | | SP1 | | | SP2 | | | SP3 | | |
| | | SPC | AO | SD | SPC | AO | SD | SPC | AO | SD |
| 8 | SPC+SD+AO | 84% | 23% | 93% | 85% | 47% | 65% | 57% | 78% | 86% |
| 9 | SPC+SD+AO (-OP) | 87% | 27% | 93% | 81% | 54% | 65% | 64% | 68% | 88% |
| 10 | SPC+SD+AO(MMNL) | 89% | 35% | 90% | 81% | 74% | 64% | 55% | 93% | 94% |
| 11 | SPC+SD+AO(-CoT) | 63% | 45% | 31% | 75% | 32% | 16% | 66% | 47% | 20% |

B.3. Aggregated choice distribution

SP1 A: Natural gas boiler B: Hydrogen-ready boiler C: Air-source heat pump
SP2 A: No retrofit B: Minor retrofit C: Major retrofit
SP3 A: Full ownership B: Minor retrofit C: Service-based

**Table B7. Choice distribution of LLaMA 3.1**

| | | SP1 | | | SP2 | | | SP3 | | |
|---|---|---|---|---|---|---|---|---|---|---|
| | | A | B | C | A | B | C | A | B | C |
| | Heating SP Survey | 173 | 194 | 194 | 236 | 281 | 44 | 253 | 167 | 141 |
| 1 | N (default) | 16 | 294 | 223 | 21 | 447 | 65 | 319 | 85 | 133 |
| 2 | SPC | 83 | 266 | 190 | 58 | 385 | 89 | 306 | 135 | 91.4 |
| 3 | SD | 25 | 270 | 238 | 35 | 400 | 98 | 237 | 153 | 138 |
| 4 | AO | 4 | 321 | 204 | 17 | 174 | 323 | 197 | 114 | 201 |
| 5 | SPC+SD | 120 | 180 | 233 | 39 | 409 | 80 | 237 | 167 | 122 |
| 6 | SD+AO | 8 | 282 | 237 | 25 | 291 | 207 | 186 | 114 | 213 |
| 7 | SPC+AO | 26 | 208 | 301 | 21 | 314 | 194 | 215 | 147 | 165 |
| 8 | SPC+SD+AO | 50 | 150 | 325 | 20 | 328 | 174 | 187 | 195 | 146 |
| 9 | SPC+SD+AO (-OP) | 44 | 155 | 324 | 23 | 326 | 175 | 191 | 195 | 140 |
| 10 | SPC+SD+AO (MMNL) | 94 | 157 | 286 | 82 | 382 | 76 | 218 | 171 | 151 |
| 11 | SPC+SD+AO (-CoT) | 36 | 137 | 341 | 10 | 380 | 87 | 73.2 | 273 | 151 |

**Table B8. Jensen–Shannon distance for LLaMA 3.1 choice simulation**

| | | SP1 | SP2 | SP3 | Average |
|---|---|---|---|---|---|
| 1 | N (default) | 0.34 | 0.41 | 0.15 | 0.30 |
| 2 | SPC | 0.17 | 0.31 | 0.11 | 0.20 |
| 3 | SD | 0.31 | 0.37 | 0.01 | 0.23 |
| 4 | AO | 0.40 | 0.57 | 0.13 | 0.37 |
| 5 | SPC+SD | 0.09 | 0.36 | 0.02 | 0.16 |
| 6 | SD+AO | 0.37 | 0.46 | 0.15 | 0.32 |
| 7 | SPC+ AO | 0.31 | 0.46 | 0.06 | 0.28 |
| 8 | SPC+SD+AO | 0.27 | 0.45 | 0.09 | 0.27 |
| 9 | SPC+SD+AO (-OP) | 0.28 | 0.44 | 0.08 | 0.27 |
| 10 | SPC+SD+AO (MMNL) | 0.17 | 0.26 | 0.04 | 0.16 |
| 11 | SPC+SD+AO (-CoT) | 0.31 | 0.45 | 0.30 | 0.35 |

**Table B9. Choice distribution of Mistral**

| | | SP1 | | | SP2 | | | SP3 | | |
|---|---|---|---|---|---|---|---|---|---|---|
| | | A | B | C | A | B | C | A | B | C |
| | Heating SP Survey | 173 | 194 | 194 | 236 | 281 | 44 | 253 | 167 | 141 |
| 1 | N (default) | 50 | 285 | 219 | 6 | 461 | 81 | 258 | 124 | 169 |
| 2 | SPC | 55 | 365 | 118 | 0 | 508 | 36 | 250 | 109 | 187 |
| 3 | SD | 207 | 143 | 205 | 24 | 475 | 47 | 141 | 295 | 120 |
| 4 | AO | 2 | 263 | 288 | 2 | 61 | 480 | 54 | 436 | 59 |
| 5 | SPC+SD | 201 | 255 | 89 | 14 | 519 | 6 | 119 | 222 | 205 |
| 6 | SD+AO | 114 | 217 | 203 | 5 | 312 | 222 | 88 | 338 | 129 |
| 7 | SPC+AO | 9 | 297 | 230 | 10 | 401 | 132 | 94 | 260 | 187 |
| 8 | SPC+SD+AO | 73 | 271 | 200 | 17 | 446 | 89 | 95 | 304 | 150 |
| 9 | SPC+SD+AO (-OP) | 76 | 275 | 194 | 10 | 451 | 87 | 91 | 284 | 178 |
| 10 | SPC+SD+AO (MMNL) | 44 | 376 | 112 | 6 | 476 | 66 | 70 | 300 | 176 |
| 11 | SPC+SD+AO (-CoT) | 43 | 256 | 250 | 39 | 443 | 69 | 49 | 360 | 146 |

**Table B10. Jensen–Shannon distance for Mistral choice simulation**

| | | SP1 | SP2 | SP3 | Average |
|---|---|---|---|---|---|
| 1 | N (default) | 0.24 | 0.47 | 0.08 | 0.26 |
| 2 | SPC | 0.30 | 0.51 | 0.11 | 0.31 |
| 3 | SD | 0.08 | 0.40 | 0.21 | 0.23 |
| 4 | AO | 0.40 | 0.76 | 0.44 | 0.53 |
| 5 | SPC+SD | 0.18 | 0.48 | 0.21 | 0.29 |
| 6 | SD+AO | 0.09 | 0.52 | 0.30 | 0.31 |
| 7 | SPC+AO | 0.37 | 0.46 | 0.26 | 0.36 |
| 8 | SPC+SD+AO | 0.19 | 0.43 | 0.27 | 0.30 |
| 9 | SPC+SD+AO (-OP) | 0.19 | 0.45 | 0.27 | 0.30 |
| 10 | SPC+SD+AO (MMNL) | 0.33 | 0.47 | 0.32 | 0.37 |
| 11 | SPC+SD+AO (-CoT) | 0.25 | 0.36 | 0.38 | 0.33 |

**Table B11. Choice distribution of GPT-3.5**

| | | SP1 | | | SP2 | | | SP3 | | |
|---|---|---|---|---|---|---|---|---|---|---|
| | | A | B | C | A | B | C | A | B | C |
| | Heating SP Survey | 173 | 194 | 194 | 236 | 281 | 44 | 253 | 167 | 141 |
| 1 | N (default) | 0 | 327 | 234 | 1 | 445 | 115 | 135 | 412 | 14 |
| 2 | SPC | 2 | 306 | 253 | 12 | 479 | 70 | 378 | 132 | 51 |
| 3 | SD | 0 | 331 | 230 | 4 | 516 | 41 | 97 | 449 | 15 |
| 4 | AO | 0 | 355 | 206 | 4 | 171 | 386 | 379 | 162 | 20 |
| 5 | SPC+SD | 1 | 302 | 258 | 4 | 488 | 69 | 341 | 159 | 61 |
| 6 | SD+AO | 0 | 338 | 223 | 4 | 416 | 141 | 373 | 180 | 8 |
| 7 | SPC+AO | 0 | 260 | 301 | 4 | 271 | 286 | 368 | 133 | 60 |
| 8 | SPC+SD+AO | 0 | 278 | 283 | 11 | 402 | 148 | 411 | 110 | 40 |
| 9 | SPC+SD+AO (-OP) | 0 | 255 | 306 | 17 | 513 | 31 | 357 | 128 | 76 |
| 10 | SPC+SD+AO (MMNL) | 0 | 324 | 237 | 5 | 501 | 55 | 370 | 160 | 31 |
| 11 | SPC+SD+AO (-CoT) | 4 | 312 | 245 | 87 | 427 | 47 | 161 | 266 | 134 |

**Table B12. Jensen–Shannon distance for GPT-3.5 choice simulation**

| | | SP1 | SP2 | SP3 | Average |
|---|---|---|---|---|---|
| 1 | N (default) | 0.42 | 0.50 | 0.41 | 0.44 |
| 2 | SPC | 0.40 | 0.44 | 0.22 | 0.35 |
| 3 | SD | 0.42 | 0.48 | 0.46 | 0.45 |
| 4 | AO | 0.43 | 0.64 | 0.29 | 0.45 |
| 5 | SPC+SD | 0.41 | 0.48 | 0.17 | 0.35 |
| 6 | SD+AO | 0.43 | 0.49 | 0.33 | 0.41 |
| 7 | SPC+AO | 0.42 | 0.56 | 0.19 | 0.39 |
| 8 | SPC+SD+AO | 0.42 | 0.46 | 0.27 | 0.38 |
| 9 | SPC+SD+AO (-OP) | 0.42 | 0.44 | 0.17 | 0.34 |
| 10 | SPC+SD+AO (MMNL) | 0.42 | 0.47 | 0.25 | 0.38 |
| 11 | SPC+SD+AO (-CoT) | 0.39 | 0.26 | 0.17 | 0.27 |

**Table B13. Choice distribution of DeepSeek-R1**

| | | SP1 | | | SP2 | | | SP3 | | |
|---|---|---|---|---|---|---|---|---|---|---|
| | | A | B | C | A | B | C | A | B | C |
| | Heating SP Survey | 173 | 194 | 194 | 236 | 281 | 44 | 253 | 167 | 141 |
| 1 | N (default) | 6 | 240 | 315 | 5 | 556 | 0 | 18 | 361 | 181 |
| 2 | SPC | 121 | 199 | 241 | 199 | 335 | 26 | 228 | 233 | 100 |
| 3 | SD | 37 | 209 | 315 | 9 | 549 | 3 | 50 | 309 | 202 |
| 4 | AO | 10 | 168 | 383 | 21 | 489 | 50 | 42 | 292 | 227 |
| 5 | SPC+SD | 114 | 200 | 247 | 190 | 347 | 24 | 216 | 249 | 96 |
| 6 | SD+AO | 41 | 189 | 331 | 9 | 541 | 11 | 62 | 243 | 256 |
| 7 | SPC+AO | 59 | 209 | 293 | 169 | 366 | 26 | 221 | 221 | 119 |
| 8 | SPC+SD+AO | 111 | 180 | 270 | 177 | 363 | 21 | 226 | 213 | 122 |
| 9 | SPC+SD+AO (-OP) | 107 | 200 | 254 | 175 | 365 | 21 | 229 | 185 | 147 |
| 10 | SPC+SD+AO (MMNL) | 99 | 200 | 262 | 175 | 358 | 28 | 236 | 215 | 110 |

**Table B14. Jensen–Shannon distance for DeepSeek-R1 choice simulation**

| | | SP1 | SP2 | SP3 | Average |
|---|---|---|---|---|---|
| 1 | N (default) | 0.38 | 0.53 | 0.45 | 0.46 |
| 2 | SPC | 0.10 | 0.09 | 0.11 | 0.10 |
| 3 | SD | 0.28 | 0.50 | 0.36 | 0.38 |
| 4 | AO | 0.39 | 0.42 | 0.38 | 0.40 |
| 5 | SPC+SD | 0.11 | 0.11 | 0.13 | 0.12 |
| 6 | SD+AO | 0.28 | 0.49 | 0.33 | 0.37 |
| 7 | SPC+AO | 0.23 | 0.13 | 0.09 | 0.15 |
| 8 | SPC+SD+AO | 0.13 | 0.13 | 0.07 | 0.11 |
| 9 | SPC+SD+AO (-OP) | 0.12 | 0.14 | 0.04 | 0.10 |
| 10 | SPC+SD+AO (MMNL) | 0.14 | 0.12 | 0.08 | 0.11 |

B.4. Stratified analysis

**Table B15. Scenario and subscenario**

| Scenario ID | Component and stratified group |
|---|---|
| 8 | SPC+SD+AO |
| S1 | Child-Free [Size: 369] |
| S2 | With-Children [Size: 192] |
| S3 | No-Seniors [Size: 385] |
| S4 | With-Seniors [Size: 176] |
| S5 | Low-income [Size: 103] |
| S6 | High-income [Size: 388] |

**Table B16. Stratified analysis**

| LLaMA 3.1 | | | | | | |
|---|---|---|---|---|---|---|
| Scenario | SP1 | | SP2 | | SP3 | |
| | Acc | F1 | Acc | F1 | Acc | F1 |
| 8 | 0.39±0.018 | 0.26 | 0.36±0.0178 | 0.20 | 0.39±0.0182 | 0.28 |
| S1 | 0.38±0.0222 | 0.27 | 0.36±0.022 | 0.19 | 0.39±0.0222 | 0.29 |
| S2 | 0.4±0.0312 | 0.26 | 0.35±0.0297 | 0.21 | 0.39±0.0307 | 0.29 |
| S3 | 0.4±0.0218 | 0.27 | 0.37±0.0216 | 0.20 | 0.39±0.0216 | 0.30 |
| S4 | 0.37±0.0318 | 0.24 | 0.33±0.0307 | 0.19 | 0.39±0.0318 | 0.28 |
| S5 | 0.41±0.0427 | 0.30 | 0.35±0.0408 | 0.19 | 0.39±0.0417 | 0.29 |
| S6 | 0.38±0.0216 | 0.25 | 0.37±0.0216 | 0.20 | 0.39±0.0216 | 0.30 |
| | | | | | | |
| Mistral | | | | | | |
| Scenario | SP1 | | SP2 | | SP3 | |
| | Acc | F1 | Acc | F1 | Acc | F1 |
| 8 | 0.4±0.041 | 0.29 | 0.47±0.041 | 0.26 | 0.44±0.041 | 0.33 |
| S1 | 0.4±0.0488 | 0.29 | 0.47±0.0515 | 0.24 | 0.44±0.0501 | 0.33 |
| S2 | 0.39±0.0677 | 0.29 | 0.47±0.0729 | 0.27 | 0.43±0.0703 | 0.32 |
| S3 | 0.38±0.0494 | 0.28 | 0.47±0.0494 | 0.25 | 0.42±0.0494 | 0.31 |

| S4 | 0.44±0.0739 | 0.30 | 0.47±0.0739 | 0.36 | 0.48±0.0739 | 0.36 |
|---|---|---|---|---|---|---|
| S5 | 0.45±0.0971 | 0.41 | 0.45±0.0971 | 0.21 | 0.4±0.0971 | 0.30 |
| S6 | 0.38±0.049 | 0.28 | 0.48±0.049 | 0.26 | 0.44±0.049 | 0.33 |
| | | | | | | |
| GPT-3.5 | | | | | | |
| Scenario | SP1 | | SP2 | | SP3 | |
| | Acc | F1 | Acc | F1 | Acc | F1 |
| 8 | 0.38±0.0401 | 0.30 | 0.44±0.041 | 0.32 | 0.58±0.041 | 0.49 |
| S1 | 0.38±0.0488 | 0.31 | 0.45±0.0515 | 0.29 | 0.58±0.0501 | 0.49 |
| S2 | 0.36±0.0677 | 0.28 | 0.43±0.0677 | 0.33 | 0.57±0.0703 | 0.50 |
| S3 | 0.37±0.0481 | 0.29 | 0.48±0.0494 | 0.34 | 0.54±0.0494 | 0.46 |
| S4 | 0.4±0.0711 | 0.31 | 0.37±0.0739 | 0.25 | 0.66±0.0682 | 0.57 |
| S5 | 0.38±0.0922 | 0.29 | 0.38±0.0971 | 0.23 | 0.58±0.0971 | 0.51 |
| S6 | 0.38±0.049 | 0.30 | 0.46±0.049 | 0.33 | 0.57±0.049 | 0.48 |
| | | | | | | |
| DeepSeek-R1 | | | | | | |
| Scenario | SP1 | | SP2 | | SP3 | |
| | Acc | F1 | Acc | F1 | Acc | F1 |
| 8 | 0.64±0.0392 | 0.64 | 0.81±0.0321 | 0.73 | 0.72±0.0365 | 0.70 |
| S1 | 0.64±0.0488 | 0.64 | 0.85±0.0366 | 0.77 | 0.71±0.0474 | 0.69 |
| S2 | 0.65±0.0677 | 0.64 | 0.72±0.0625 | 0.67 | 0.74±0.0625 | 0.72 |
| S3 | 0.64±0.0481 | 0.64 | 0.77±0.0416 | 0.68 | 0.71±0.0442 | 0.70 |
| S4 | 0.65±0.0682 | 0.65 | 0.89±0.0483 | 0.85 | 0.73±0.0653 | 0.70 |
| S5 | 0.69±0.0874 | 0.69 | 0.84±0.0728 | 0.79 | 0.69±0.0874 | 0.67 |
| S6 | 0.62±0.0477 | 0.61 | 0.8±0.0399 | 0.71 | 0.71±0.0438 | 0.70 |